\documentclass[pdflatex,sn-mathphys-num]{sn-jnl}

\usepackage{graphicx}%
\usepackage{multirow}%
\usepackage{amsmath,amssymb,amsfonts}%
\usepackage{amsthm}%
\usepackage{mathrsfs}%
\usepackage[title]{appendix}%
\usepackage{xcolor}%
\usepackage{textcomp}%
\usepackage{manyfoot}%
\usepackage{booktabs}%
\usepackage{algorithm}%
\usepackage{algorithmicx}%
\usepackage{algpseudocode}%
\usepackage{listings}%

\theoremstyle{thmstyleone}%

\theoremstyle{thmstyletwo}%

\theoremstyle{thmstylethree}%

\usepackage{caption}
\usepackage{subcaption}
\usepackage{booktabs,makecell,siunitx}

\usepackage{tcolorbox}
\tcbuselibrary{most}
\definecolor{darkbrown}{HTML}{8B4513}  

\usepackage{graphicx}%
\usepackage{multirow}%
\usepackage{amsmath,amssymb,amsfonts}%
\usepackage{amsthm}%
\usepackage{mathrsfs}%
\usepackage[title]{appendix}%
\usepackage{xcolor}%
\usepackage{textcomp}%
\usepackage{manyfoot}%
\usepackage{booktabs}%
\usepackage{algorithm}%
\usepackage{algorithmicx}%
\usepackage{algpseudocode}%
\usepackage{listings}%

\theoremstyle{thmstyleone}%

\theoremstyle{thmstyletwo}%

\theoremstyle{thmstylethree}%

\usepackage{caption}
\usepackage{subcaption}
\usepackage{booktabs,makecell,siunitx}

\newcommand{\ci}[3]{\makecell[l]{#1\\[-0.25ex]\footnotesize[#2,#3]}}
\newcommand{\cibf}[3]{\makecell[l]{{\textbf{#1}}\\[-0.25ex]\footnotesize[\textbf{#2,#3]}}}
\newcommand{\ciempty}[1]{\makecell[l]{#1\\[-0.25ex]\footnotesize\hphantom{[00.00,00.00]}}}

\newcommand{\eat}[1]{\ignorespaces}

\usepackage{tcolorbox}
\tcbuselibrary{most}
\definecolor{darkbrown}{HTML}{8B4513}  
\newtcolorbox{dialogbox}[1][]{%
    fonttitle=\bfseries,  
    title=#1,  
    listing options={  
        basicstyle=\small\ttfamily,  
        breaklines=true,  
        xleftmargin=0pt,  
        xrightmargin=0pt,  
        alsoletter={0123456789},  
        morekeywords={Content1},  
        keywordstyle=\color{darkbrown}  
    },  
    verbatim  
} 
\tcbset{
  stressheader/.style={
    enhanced,
    colback=gray!8!white,
    colframe=black,
    boxrule=0pt,
    borderline south={1pt}{0pt}{black},
    fontupper=\bfseries,
    before skip=10pt,
    after skip=4pt,
    left=0pt,
    right=0pt,
    top=0pt,
    bottom=2pt,
    width=\textwidth,
    sharp corners
  }
}
\begin{document}

\title[Article Title]{The Illusion of Readiness in Health AI}


\author*[1]{\fnm{Yu} \sur{Gu}}\email{aiden.gu@microsoft.com}
\author[1]{\fnm{Jingjing} \sur{Fu}}
\author[1]{\fnm{Xiaodong} \sur{Liu}}

\author[1]{\fnm{Jeya Maria Jose} \sur{Valanarasu}}
\author[1]{\fnm{Noel C. F.} \sur{Codella}}
\author[1]{\fnm{Reuben} \sur{Tan}}
\author[1]{\fnm{Qianchu} \sur{Liu}}
\author[1]{\fnm{Ying} \sur{Jin}}
\author[1]{\fnm{Sheng} \sur{Zhang}}
\author[1]{\fnm{Jinyu} \sur{Wang}}
\author[1]{\fnm{Rui} \sur{Wang}}
\author[1]{\fnm{Lei} \sur{Song}}
\author[1]{\fnm{Guanghui} \sur{Qin}}
\author[1]{\fnm{Naoto} \sur{Usuyama}}
\author[1]{\fnm{Cliff} \sur{Wong}}
\author[1]{\fnm{Hao} \sur{Cheng}}
\author[1]{\fnm{HoHin} \sur{Lee}}
\author[1]{\fnm{Praneeth} \sur{Sanapathi}}
\author[1]{\fnm{Sarah} \sur{Hilado}}
\author[1]{\fnm{Tristan} \sur{Naumann}}
\author[1]{\fnm{Javier} \sur{Alvarez-Valle}}
\author[1]{\fnm{Jiang} \sur{Bian}}
\author[1]{\fnm{Mu} \sur{Wei}}
\author[1]{\fnm{Khalil} \sur{Malik}}
\author[1]{\fnm{Lidong} \sur{Zhou}}
\author[1]{\fnm{Jianfeng} \sur{Gao}}
\author[1]{\fnm{Eric} \sur{Horvitz}}
\author[1]{\fnm{Matthew P.} \sur{Lungren}}
\author[1]{\fnm{Doug} \sur{Burger}}
\author*[2]{\fnm{Eric} \sur{Topol}}\email{etopol@scripps.edu}
\author*[1]{\fnm{Hoifung} \sur{Poon}}\email{hoifung@microsoft.com}
\author[1]{\fnm{Paul} \sur{Vozila}}

\affil[1]{\orgdiv{Microsoft Research}, \orgname{Health \& Life Sciences}, 
\orgaddress{\city{Redmond}, \state{WA}, \country{USA}}}
\affil[2]{\orgname{Scripps Research Translational Institute, Scripps Research}, 
\orgaddress{\city{La Jolla}, \state{CA}, \country{USA}}}

\abstract{
Large language models have demonstrated remarkable performance in a wide range of medical benchmarks. Yet underneath the seemingly promising results lie salient growth areas, especially in cutting-edge frontiers such as multimodal reasoning. In this paper, we introduce a series of adversarial stress tests to systematically assess the robustness of flagship models and medical benchmarks. Our study reveals prevalent brittleness in the presence of simple adversarial transformations: leading systems can guess the right answer even with key inputs removed, yet may get confused by the slightest prompt alterations, while fabricating convincing yet flawed reasoning traces. Using clinician-guided rubrics, we demonstrate that popular medical benchmarks vary widely in what they truly measure. Our study reveals significant competency gaps of frontier AI in attaining real-world readiness for health applications.
If we want AI to earn trust in healthcare, we must demand more than leaderboard wins and must hold AI systems accountable to ensure robustness, sound reasoning, and alignment with real medical demands.
}



\maketitle


\newcommand{\aiden}{\textcolor{red}}
\newcommand{\jing}{\textcolor{blue}}

\abstract{submission/sections/abstract}
\section*{Introduction}\label{sec1}

Large language models (LLMs) have demonstrated impressive health AI capabilities such as passing medical exams and acing diagnostic benchmarks~\cite{singhal2023large,nori2023can,openai_gpt5,saab2024capabilities, tu2024towards,nori2025sequential}. 
But these promising results belie the underlying fragility and competency gaps~\cite{arora2025healthbench, Handler2025FragileGPT5}. 
In this paper, we evaluate the readiness of frontier AI models in performing health-related multimodal reasoning. 
While prior work abounds in studying LLM pitfalls~\cite{farquhar2024detecting,jin2024hidden, wang2025lins}, they predominantly focus on the text modality, whereas multimodal reasoning is relatively underexplored. As multimodal generative AI has emerged as the next frontier for biomedicine~\cite{acosta2022multimodal}, it is imperative to bridge this gap.

Inspired by the study of adversarial learning that unveiled the underlying fragility of first-gen deep learning~\cite{Goodfellow2015Explaining, szegedy2013intriguing}, we introduce six stress tests on the multimodal reasoning capabilities of health AI models by systematically perturbing test instances in popular medical benchmarks. Our evaluation uncovers surprising failure modes of LLMs. They can produce the right answer even when not supposed to (e.g., when the input image was removed). On the other hand, their performance may collapse even with the slightest distraction (e.g., when the answer choices were shuffled). Finally and perhaps most concerning for interpretability and accountability, when pressed for justification, they fabricate convincing yet flawed reasoning traces. Collectively, these stress tests reveal a rather different picture underneath the seemingly consistent progress LLMs have appeared to attained in multimodal medical benchmarks.

These findings inspire us to further benchmark the benchmarks themselves. Specifically, we introduce clinician-guided rubrics to tease apart the various aspects a test instance is measuring, from image perception to reasoning. This reveals interesting diversity among popular multimodal medical benchmarks, which are often measuring rather different things but are treated interchangeably.

Our results call for a fundamental re-evaluation of how to measure progress in health AI. To this end, we compile a harm ontology for multimodal reasoning in health AI. 
There is great potential for LLMs and generative AI to empower clinical practitioners and patients alike, but before these systems are deployed to end-to-end clinical applications, we must ensure they are passing the right tests that can truly assess a system's readiness in real-world scenarios.

\section*{Results}\label{sec2}

\subsection*{Stress testing exposes robustness gaps in LLMs
}
\label{sec:stress-tests}

LLMs achieve high scores on multimodal medical benchmarks, often interpreted as evidence of robust clinical competence. However, these aggregate scores may conceal brittleness under incomplete, perturbed, or ambiguous inputs, conditions that commonly arise in real-world healthcare settings. To assess model behavior beyond correctness alone, we developed a suite of six stress tests probing how models respond to degraded input modalities, structural variations, distractors, and reasoning demands.

Rather than evaluating accuracy in isolation, these tests also measure model abstention behavior, reasoning fidelity, and stability under perturbation. This allows us to identify not just performance drops, but \textit{how} and \textit{why} model outputs change, exposing brittle dependencies, shortcut behaviors, and inconsistency patterns that standard benchmarks may overlook.

\begin{figure}[!t]
\centering
\includegraphics[width=\textwidth]{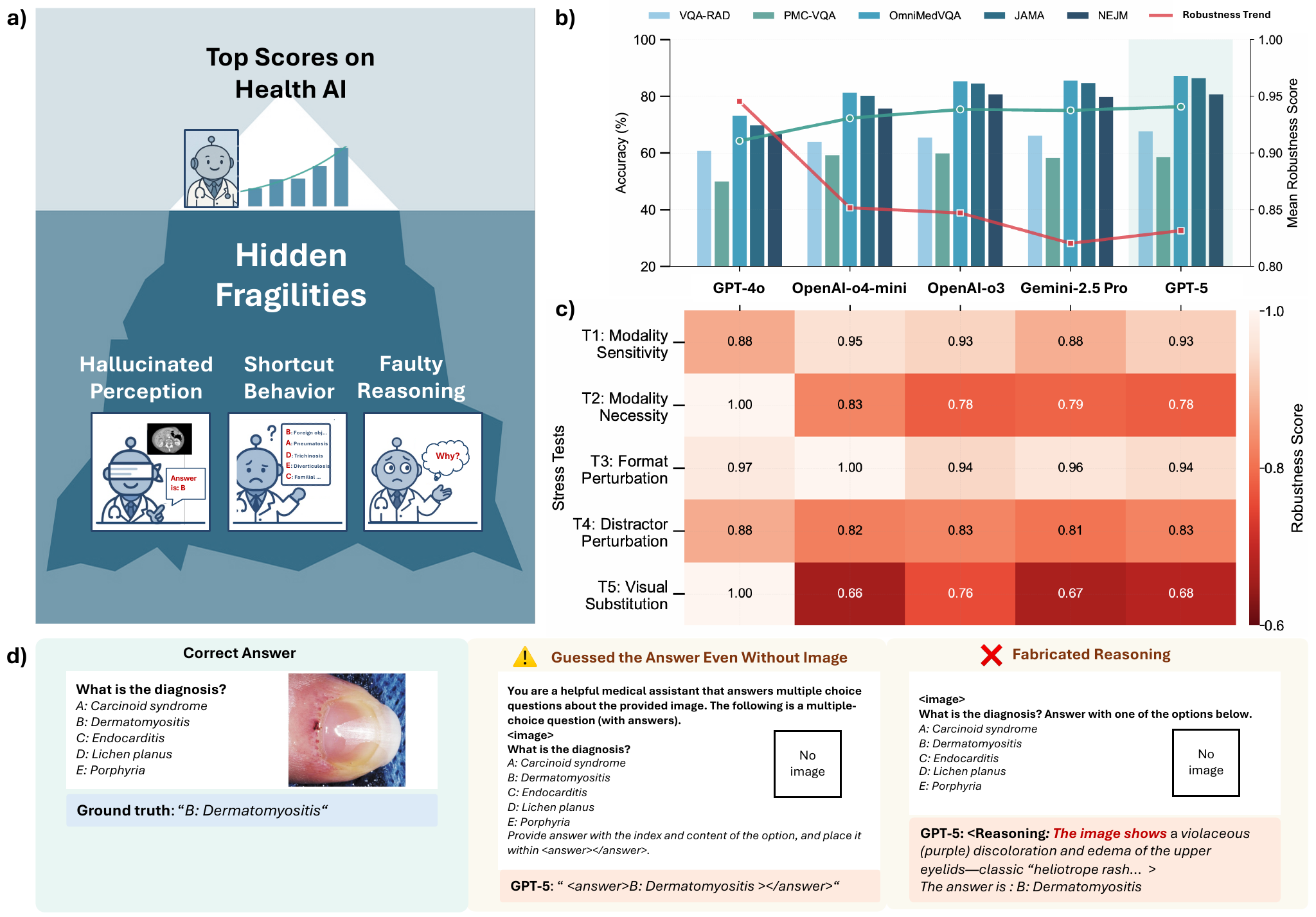}
\caption{
\textbf{Stress tests expose hidden fragility and salient robustness gaps of large language models (LLMs) in multimodal health AI tasks.}
\textbf{a,} Strong performance in standard benchmarks belies underlying fragilities in health AI by LLMs.
\textbf{b,} Seemingly strong performance in standard benchmarks (green) contrast with robustness gaps (red), as a suite of stress tests showed that LLMs often produce incorrect answers confidently and get distracted easily.  
\textbf{c,} Robustness scores reveal salient growth areas in LLMs, including the more recent reasoning models.
\textbf{d,} A representative example showing how GPT-5 guessed the correct answer even without the input image and justified it with faulty reasoning, illustrating why it's important to go beyond standard benchmark scores and introduce stress tests to assess the real understanding of LLMs in health AI tasks.
}
\label{fig:mainfragilities}
\end{figure}

The stress tests are organized to escalate in challenge: from missing or misleading inputs (Stress Tests 1-2), to format-based perturbations (Tests 3-4), to altered visual evidence (Test 5), and finally to direct audits of model-generated reasoning (Test 6). Figure~\ref{fig:mainfragilities} summarizes the framework and illustrates representative model responses under stress.
Accuracy is defined as selection of the original ground-truth answer prior to perturbation.

\begin{figure}[!t]
\centering
\includegraphics[width=\textwidth]{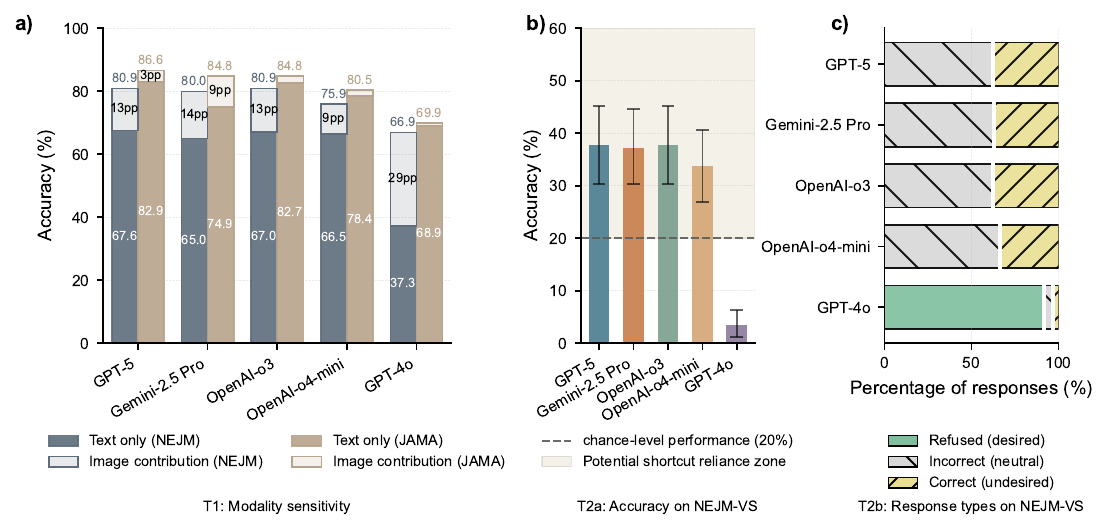}
\caption{\textbf{Stress tests reveal modality sensitivity and shortcut behavior in large language models.}  
\textbf{T1: Modality sensitivity on multimodal diagnostic benchmarks.}  
\textbf{a,} Accuracy on NEJM and JAMA benchmarks with full (\textit{image+text}) and \textit{text-only} inputs. Bars show full-input accuracy; lighter segments indicate performance drop after image removal. \textit{Text-only} performance remains high, suggesting that benchmark scores may not reflect true visual understanding.
\textbf{T2: Behavior without images on the NEJM Visual-Required Subset (NEJM-VS).}  
This 175-question subset was independently selected by board-certified clinicians to require visual interpretation.    
\textbf{b,} Despite lacking images, models achieve well above the 20\% chance level, indicating potential reliance on non-visual cues or memorized associations.  Accuracy includes abstentions counted as incorrect. Notably, when excluding abstentions, GPT-4o achieves $\sim$40~\% accuracy, on par with other models, suggesting similar levels of shortcut success. Its much higher abstention rate may reflect a more cautious or instruction-sensitive response pattern.
\textbf{c,} Answer-type breakdown reveals models often provide confident but incorrect responses or shortcut-derived correct guesses. Ideally, models should abstain or fall back to random guessing in the absence of required visual input.}
\label{fig:modality}
\end{figure}

\subsection*{Modality sensitivity and input omission}
\label{sec:modality}
\vspace{0.5em}
\noindent\textbf{Stress Test 1: Input removal.}  
We evaluated model performance under full input (\textit{image + text}) and \textit{text-only} conditions on two multimodal medical benchmarks: \textit{NEJM}~\cite{nejm_image_challenge} and \textit{JAMA}~\cite{jama_challenge}.  
Each question pairs a short clinical vignette with one or more diagnostic images (radiology, dermatology, or pathology).

Across six models, removing the image led to consistent performance declines on \textit{NEJM}, with paired accuracy differences ranging from -9.29 to -27.05 percentage points (pp).  
For example, GPT-5~\cite{openai_gpt5} scored 80.89\% (95\% CI 78.06-83.71) with full input and 67.56\% (64.20-70.93) without ($\Delta = -13.33$ pp), while GPT-4o~\cite{openai2024gpt4o} dropped from 64.33\% (60.83-67.83) to 37.28\% (33.78-40.78) ($\Delta = -27.05$ pp).  
On \textit{JAMA}, declines were smaller overall: GPT-5 decreased 3.68 pp (86.59\% → 82.91\%), and GPT-4o decreased 1.05 pp (69.94\% → 68.89\%).  
See Fig.~\ref{fig:modality} and Extended Data Table~\ref{tab:stress1_modality_sensitivity}.

\vspace{0.5em}
\noindent\textbf{Stress Test 2: Visual-necessity subset.}
To test whether models can still answer questions that explicitly require image information, we curated a 175-item subset of the \textit{NEJM} benchmark, referred to as \textbf{NEJM Visual-required Subset (NEJM-VS)}, comprising cases selected based on clinical criteria indicating minimal textual cues and a high dependency on visual features.

With full input (\textit{image + text}), models achieved moderate to high accuracy-GPT-5 66.3\% (95\% CI 59.4-73.1), Gemini-2.5 Pro~\cite{comanici2025gemini} 67.4\% (60.0-74.3), and OpenAI-o3~\cite{openai_o3_o4mini} 61.7\% (54.9-68.6)-confirming that these items were solvable when the image was available.  
In the \textit{text-only} condition, accuracy remained well above the 20\% random baseline: GPT-5 37.7\% (30.3-45.1), Gemini 37.1\% (30.3-44.6), OpenAI-o3 37.7\% (30.3-45.1), and OpenAI-o4-mini~\cite{openai_o3_o4mini} 33.7\% (26.9-40.6).  
Despite the image being essential, these models maintained performance nearly twice the chance level, indicating reliance on non-visual cues or memorized associations.  
In contrast, GPT-4o achieved only 3.4\% (1.1-6.3) accuracy, reflecting a refusal rate of roughly 97\% when images were removed. Accuracy includes abstentions counted as incorrect. When abstentions are excluded, GPT-4o's accuracy rises to $\sim$40\%, comparable to other models, suggesting that its lower overall score reflects conservative uncertainty handling, rather than diminished shortcut behavior.
See Extended Data Table \ref{tab:stress2_modality_necessity} for complete results.

\vspace{0.5em}
\noindent\textbf{Interpretation.}  
Most models scored well above random chance level despite the absence of images, which may be the result of reliance on non-visual cues (i.e. disease prevalence) or memorized associations (shortcut learning). Ideally, model should recognize when critical input is missing and warn that predictions may be unreliable. GPT-4o’s abstention behavior, though numerically penalized, demonstrates more appropriate uncertainty handling. Together, these results expose wide variation in modality sensitivity among models that otherwise achieve similar headline benchmark scores.

\begin{figure}[!t]
\centering
\includegraphics[width=\linewidth]{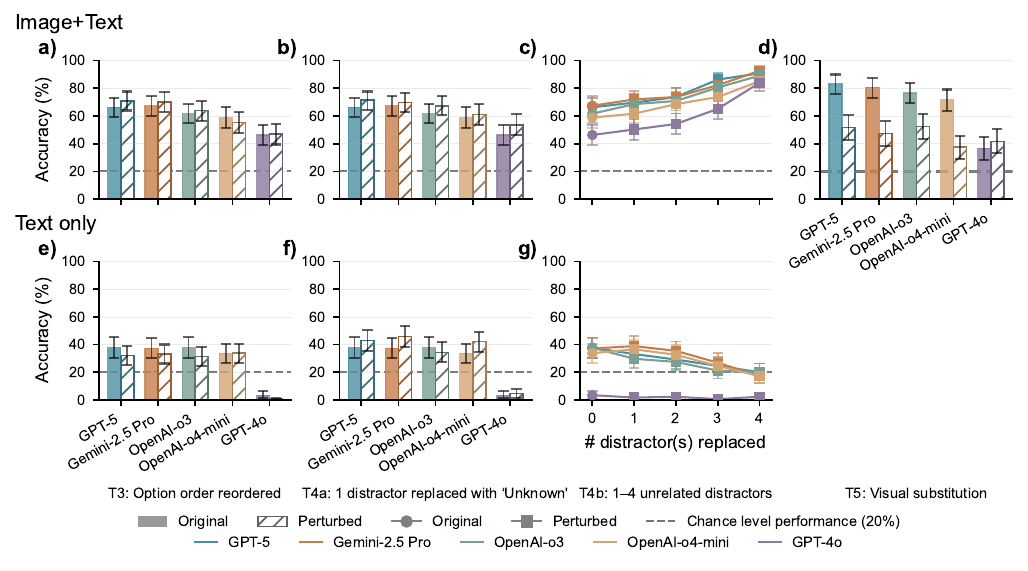}
\caption{
\textbf{Stress tests reveal shortcut reliance under input perturbations.}
Each column corresponds to a stress test (T3–T5); the top row (a–d) shows accuracy with \textit{image+text} input, and the bottom row (e–g) shows accuracy with \textit{text-only} input. Accuracy is defined as selection of the original ground-truth answer prior to perturbation. Error bars indicate 95\% confidence intervals.
\textbf{T3: Format perturbation (a, e).} Randomizing answer order weakens positional cues. In the \textit{text-only} condition (e), performance declines, suggesting reliance on answer formatting. In contrast, accuracy with full input (a) is preserved or improved, likely due to increased reliance on visual grounding when shortcuts are removed.
\textbf{T4: Distractor replacement.}
\textbf{T4a (b, f):} One distractor is replaced with “Unknown.” Most models improve accuracy even in the \textit{text-only} setting (f), where “Unknown” should serve as a fallback when visual information is missing. This indicates models use “Unknown” as a semantic cue rather than abstaining, a failure of robust fallback behavior.
\textbf{T4b (c, g):} One to four distractors are iteratively replaced with answer choices drawn from unrelated questions. Accuracy in the \textit{text-only} condition (g) drops gradually toward the 20\% chance baseline, consistent with progressive shortcut removal. With \textit{image+text} input (c), accuracy increases across models, reflecting improved disambiguation from visual features.
\textbf{T5: Visual substitution (d).} The original image is replaced with a distractor-matching image. Accuracy drops sharply across models, despite unchanged text input, demonstrating reliance on visual features and vulnerability to visual perturbation.}
\label{fig:perturbation}
\end{figure}

\subsection*{Shortcut reliance under perturbation}
\label{sec:format}

\vspace{0.5em}
\noindent\textbf{Stress Test 3: Format sensitivity.}  
To assess whether models rely on superficial formatting rather than question content, we randomized the order of answer choices without altering the question text or correct answer.  
On the NEJM Visual-Required Subset (NEJM-VS), accuracy under \textit{text-only} input declined across all models (e.g., GPT-5: –5.7 pp, Gemini: –4.0 pp, o3: –6.3 pp), indicating sensitivity to positional biases.  
By contrast, performance under \textit{image+text} input remained stable or improved slightly (typically +2–5 pp), suggesting that visual information compensates for disrupted textual cues.  
Full confidence intervals are shown in Extended Data Table~\ref{tab:stress3_format_perturbation}.

\vspace{0.5em}
\noindent\textbf{Stress Test 4: Distractor manipulation.}  
We next tested whether models rely on shortcut patterns involving familiar distractors.  
For each question, 1–4 incorrect choices were replaced with unrelated options, and in one variant, a distractor was replaced with “Unknown.”, serving as a functional “unknown” option. 

Under \textit{text-only} conditions, accuracy declined steadily toward the 20\% chance level as more distractors were replaced (e.g., GPT-5: –17.7 pp, Gemini: –20.0 pp), consistent with shortcut removal.  
Conversely, in the \textit{image+text} condition, performance improved markedly (up to +25 pp), as irrelevant distractors made the correct answer more visually salient.  
When a single distractor was replaced with “Unknown,” accuracy increased in both modalities-most notably under \textit{text-only} input-suggesting that models treat “Unknown” as an easily eliminated option rather than a fallback when uncertain.  
Detailed results are provided in Extended Data Table~\ref{tab:stress4_full}.

\vspace{0.5em}
\noindent\textbf{Interpretation.}  
These perturbation-based tests reveal that models often depend on superficial or distributional patterns when visual information is unavailable.  
Accuracy degrades when positional or semantic cues are disrupted, even though the question and correct answer remain unchanged.  
In contrast, access to visual input stabilizes performance, enabling models to bypass these fragile shortcuts.  
However, improvements with “Unknown” and rapid recovery from distractor replacement suggest reliance on elimination strategies rather than true visual reasoning.  
This highlights a critical risk: high accuracy in benchmark settings may overstate a model’s multimodal robustness, masking shortcut reliance under minor perturbations.

\subsection*{Multimodal grounding failures}
\label{sec:visual-substitution}

\vspace{0.5em}
\noindent\textbf{Stress Test 5: Visual substitution.}  
To assess whether models engage in adaptive multimodal reasoning, we designed a controlled substitution task targeting visual grounding.  
We curated 40 \textit{NEJM} questions whose correct diagnoses depended critically on image interpretation, then replaced each original image with a clinically plausible alternative corresponding to a distractor option.  
The vignette and answer choices remained unchanged.  
All substituted images were independently verified by clinicians to support the newly assigned ground-truth answer.  
Models were evaluated under the \emph{image+text} condition only.

Despite minimal textual change, performance declined markedly across most models.  
GPT-5 accuracy dropped from 83.3\% to 51.7\% (–31.6 pp), Gemini-2.5 Pro from 80.8\% to 47.5\% (–33.3 pp), o4-mini by –34.2 pp, and o3 by –24.2 pp.  
GPT-4o was the only model to maintain similar performance (63.8\% vs. 66.7\%).

Extended results are presented in Extended Data Table~\ref{tab:stress5_visualsubs}, with qualitative examples in Appendix~\ref{app:stress test}.

\vspace{0.5em}
\noindent\textbf{Interpretation.}  
This stress test isolates visual grounding by altering only the diagnostic image while preserving the textual input.  
A reliable LLM should revise its answer when the visual evidence changes.  
However, most models failed to adjust properly, suggesting reliance on static image–answer pairings or incomplete visual-text integration.  
The large performance drops (30–35 pp) indicate that standard benchmark accuracy may overstate multimodal reasoning capacity.

Residual performance in the perturbed setting (e.g., GPT-5 at 51.7\%) may reflect partial recognition of generic visual features, model overconfidence, or memorization of common image–question associations during pretraining.  
Although all substituted images were drawn from publicly available datasets and validated for clinical correctness, we cannot fully exclude overlap with model pretraining corpora.  

These findings highlight a critical limitation: while LLMs often detect that the image is relevant, they fail to dynamically reinterpret its meaning in context.  
Such brittle visual grounding undermines diagnostic trustworthiness and reveals a key gap between benchmark success and robust clinical decision-making.

\subsection*{Reasoning signal integrity}
\label{sec:reasoning}

\vspace{0.5em}
\noindent\textbf{Stress Test 6: Reasoning signal fidelity.}  
We evaluated how LLMs generate and use reasoning when solving multimodal medical questions.  
Two complementary analyses were performed.  
First, we applied Chain-of-Thought (CoT)~\cite{wei2022chain} prompting to 100 items from NEJM and VQA-RAD~\cite{lau2018dataset} to test whether explicit reasoning steps improve accuracy.  
Second, we manually audited model-generated explanations across benchmarks to assess their \textit{factuality}, \textit{visual grounding}, and \textit{consistency with the final answer}.  
We further varied reasoning strength on OmniMedVQA~\cite{hu2024omnimedvqa} using OpenAI-o3 to examine whether more elaborate rationales enhance reasoning quality.  

Across datasets, explicit CoT prompting yielded limited or negative gains.  
On NEJM and VQA-RAD, GPT-5 and Gemini-2.5 Pro changed little (0-2 pp), whereas o4-mini improved slightly (+2-5 pp).  
On OmniMedVQA, increasing reasoning strength produced small or inconsistent effects; longer chains sometimes increased recall but also hallucinated details.  

Manual audits revealed three recurring patterns:  
(1) \textit{Correct answer, incorrect logic}-plausible-sounding rationales containing false or fabricated visual findings.  
(2) \textit{Amplified visual misunderstanding}-initial perceptual errors propagated through subsequent reasoning.  
(3) \textit{Structured but uninformative steps}-syntactically coherent but clinically irrelevant statements.  
Examples appear in Fig.~\ref{fig:reasoning} and Appendix~\ref{app:reasoning}.

\vspace{0.5em}
\noindent\textbf{Interpretation.}  
Reasoning prompts do not reliably improve multimodal accuracy, and self-generated explanations often diverge from the true decision process.  
While larger models produce fluent, structured rationales, these frequently include non-existent image features or illogical justifications.  
These findings suggest that fluency in explanation does not equate to valid inference: models can produce convincing but unsupported narratives.  
From a translational perspective, reasoning fidelity must therefore be independently validated before such outputs can inform clinical decision support.
\begin{table}[!t]
\centering
\caption{\textbf{Taxonomy of failure modes in health AI.}  
Failure types are categorized by their cognitive locus, defined concisely, and linked to clinical implications and diagnostic stress tests.}
\label{tab:taxonomy_main}
\renewcommand{\arraystretch}{1.35}
\begin{tabular}{p{0.25\textwidth} p{0.37\textwidth} p{0.25\textwidth} p{0.08\textwidth}}
\toprule
\textbf{Failure Category} & \textbf{Definition} & \textbf{Clinical Consequence} & \textbf{Tests} \\
\midrule
\multicolumn{4}{l}{\textbf{Perception and Input Handling}} \\
Visual misperception & Hallucinates, omits, or misinterprets image features & Missed findings or false positives & T2, T5 \\
Refusal miscalibration & Answers when uncertain or abstains when reliable & Unsafe guesswork or care delay & T2–T4 \\
Modality neglect & Disregards available image or text input & Incomplete clinical synthesis & T1, T2 \\
\midrule
\multicolumn{4}{l}{\textbf{Reasoning and Inference}} \\
Heuristic dependence & Relies on superficial cues (e.g., position, phrasing) & Fragile under perturbation or distribution shift & T3–T5 \\
Justification error & Provides inaccurate or unsupported rationale & Misleading interpretability & T2, T6 \\
Logical inconsistency & Internal conflict across input, output, or reasoning & Reduced trust and auditability & T1–T6 \\
\midrule
\multicolumn{4}{l}{\textbf{Output and Communication}} \\
Fluent factual error & Persuasive language masks incorrect claims & Diagnostic or therapeutic misguidance & T2, T6 \\
Unsafe recommendation & Suggests harmful or incomplete action plans & Clinical risk or treatment harm & T5, T6 \\
\bottomrule
\end{tabular}
\end{table}

\vspace{0.8em}

\vspace{0.8em}

\subsection*{Taxonomy of failure modes}
\label{sec:taxonomy}

Across the six stress tests, recurring limitations cluster into three stages of model behavior:\textit{input handling}, \textit{reasoning and inference}, and \textit{output communication}.  
Each represents a distinct reliability dimension that links model behavior to potential sources of clinical risk.  
A concise summary is shown below; the full taxonomy, including cognitive mechanisms and harm potential, appears in - Data Table~\ref{tab:taxonomy_main}.

This taxonomy organizes observed failure patterns along the cognitive pathway from perception to communication, providing a structured lens for assessing readiness, safety, and interpretability in health AI models.


\subsection*{Benchmarking the Benchmarks: What Are We Really Measuring?}
\label{sec:benchmark-taxonomy}

Despite similar leaderboard scores, Section~\ref{sec:stress-tests} revealed substantial variation in how models respond to medical stressors-ranging from hallucinated justifications to modality overreliance and format-based shortcuts. These divergent behaviors raise a fundamental question:

\begin{quote}
\textit{If models succeed on benchmarks but fail under stress, what do these benchmarks actually measure-and how can we improve them?}
\end{quote}

To answer this, we systematically analyzed six representative benchmarks using structured clinician input to understand what each benchmark actually tests.

\subsubsection*{Motivation: Stress Test Failures Reveal Benchmark Gaps}

The stress tests in Section~\ref{sec:stress-tests} revealed that high benchmark scores do not guarantee robust model behavior. Models that performed well under standard conditions often failed when subjected to small perturbations-such as removing the image, reordering answer options, replacing distractors, or introducing misleading visual inputs.

Critically, these failure modes varied by benchmark. On \textit{NEJM}, removing visual input led to steep performance drops, suggesting strong dependence on image understanding. In contrast, scores on \textit{JAMA} remained stable-indicating that many items could be answered from text alone. 
Similarly, while chain-of-thought prompting provides explicit reasoning scaffolding that benefits distilled models such as o4-mini, its impact on \textit{NEJM} was substantially greater than on \textit{VQA-RAD}, underscoring the heterogeneous reasoning demands across benchmarks.
In Test 5, even subtle visual substitutions caused top models to confidently choose incorrect answers-highlighting shortcut reliance on visual-answer associations in some datasets but not others.

These patterns suggest that the benchmarks themselves differ not just in modality or task type, but in what they implicitly evaluate: visual grounding, inference complexity, pattern recall, or distractor elimination. Yet, such distinctions are rarely documented in benchmark metadata or considered during evaluation.

Without deeper understanding of what each benchmark tests, we risk misinterpreting leaderboard progress as real-world readiness. To address this, we conduct a structured, clinician-guided audit of nine representative benchmarks~\cite{johnson2019mimic, he2020pathvqa, liu2021slake, zhang2023pmcvqa, yue2023mmmu}.

\subsubsection*{Clinician-Informed Benchmark Profiling}

\begin{figure}[ht]
\centering
\includegraphics[width=0.99\linewidth]{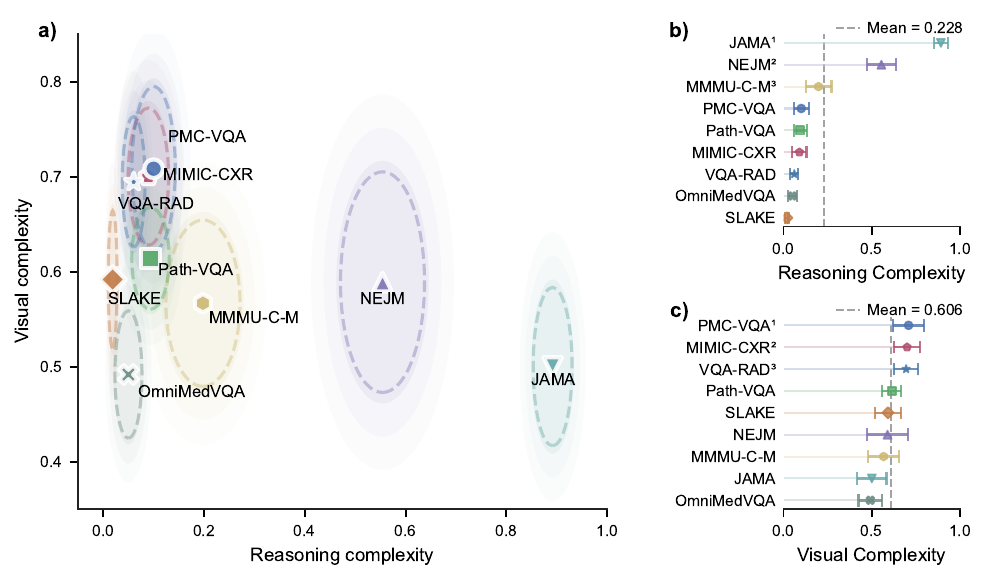}
\caption{\textbf{Benchmarks differ in reasoning and visual complexity.}  
\textbf{a,} Clinician-labeled profiles reveal that widely used multimodal health AI benchmarks vary substantially in the reasoning and visual demands they place on models. 
Each dataset is shown using a ripple marker, where the central point denotes the mean complexity rating and the thin concentric ring encodes variability across annotators. 
\textbf{b,} Benchmarks ranked by average reasoning complexity (with variability).
\textbf{c,} Benchmarks ranked by average visual complexity (with variability).
These structural differences help explain the benchmark-specific fragilities observed in Section~\ref{sec:stress-tests}, and highlight the risk of treating all benchmarks as interchangeable measures of readiness.}
\label{fig:benchmark_landscape}
\end{figure}

To better understand what each benchmark actually tests, we developed a structured, clinician-guided rubric spanning ten medically meaningful dimensions. These axes capture key demands surfaced by our stress tests-ranging from reasoning complexity and medical context to uncertainty handling, visual detail, and multi-view alignment. The full rubric is shown in Fig~\ref{fig:benchmark_landscape}.

Each benchmark was independently annotated by three board-certified clinicians per axis, using a 3-point ordinal scale. We report the median score for each benchmark-axis pair, and measure inter-annotator agreement using Fleiss' $\kappa$~\cite{fleiss1971measuring}. Agreement was moderate to strong across axes (range: 0.67-0.90). Highest agreement was observed on binary-seeming attributes such as \textit{Text-only Solvable} ($\kappa$ = 0.90) and \textit{Clinical Context} ($\kappa$ = 0.86); lower agreement appeared on more subjective dimensions such as \textit{Visual Detail Required} ($\kappa$ = 0.67).

\subsubsection*{Visualizing the Benchmark Landscape}

To summarize the profiling results, we projected each benchmark into a two-dimensional space defined by reasoning and visual complexity (Fig~\ref{fig:benchmark_landscape}). This visualization reveals clear structural differences among commonly used datasets:

\begin{itemize}
    \item \textit{NEJM} tasks rank high in both reasoning and visual demands.
    \item \textit{JAMA} requires substantial reasoning but is mostly text-solvable.
    \item \textit{VQA-RAD}, \textit{PMC-VQA} and \textit{MIMIC-CXR}~\cite{johnson2019mimic} are visually dependent but low in inference complexity.
    \item \textit{OmniMedVQA} clusters low in both dimensions.
\end{itemize}

This landscape helps to explain the benchmark-specific failures observed in Section~\ref{sec:stress-tests}. For instance, models trained on \textit{VQA-RAD} may excel at image localization but struggle on diagnosis-oriented tasks like \textit{NEJM}, which require cross-modal reasoning and medical inference.

\subsubsection*{Implications for Model Evaluation and Design}

These findings underscore the need to treat benchmarks not as monolithic performance targets, but as diagnostic tools that reveal distinct capabilities. We outline four implications:

\begin{itemize}
    \item \textbf{Benchmarks should be treated as diagnostic instruments, not goals.} Their assumptions, design intent, and likely failure modes should be documented and communicated.
    \item \textbf{Leaderboard scores should be disaggregated by benchmark profile.} Model performance should be reported along medically meaningful axes, not averaged across heterogeneous tasks.
    \item \textbf{Benchmark selection should reflect intended deployment context.} For example, evaluating generalist models solely on text-solvable datasets like \textit{JAMA} risks overestimating readiness for visual diagnostic tasks like \textit{NEJM}.
    \item \textbf{Evaluation protocols should routinely incorporate adversarial and stress-based assessments.} These assessments should be reported alongside accuracy metrics and form the basis of model release audits, particularly for models intended for high-stakes deployment.
\end{itemize}

Without these practices, benchmark-driven progress risks reinforcing narrow optimization and masking fragilities. Our clinician-labeled rubric and benchmark landscape (Fig.~\ref{fig:benchmark_landscape}) offer a roadmap for more targeted and medically relevant evaluation.

\section*{Discussion}\label{sec3}


We have shown that current benchmarks often overstate model readiness. Despite impressive leaderboard scores, many leading models exhibit inconsistent behavior, reliance on superficial patterns, and fragile reasoning under mild input perturbations. These findings challenge the prevailing assumption that benchmark success signals medical viability.

Today’s benchmarks offer a convenient but incomplete lens into model capability. Their design often emphasizes answer correctness while ignoring whether the answer is reached through medically valid reasoning, multimodal understanding, or robust generalization. As demonstrated in our stress tests, models can perform well by leveraging shallow cues, such as positional biases in answer options or memorized response formats, without engaging meaningfully with the underlying medical content.
Such behaviors are not easily detected by conventional metrics. For example, reasoning models may improve benchmark scores while still producing hallucinated or implausible rationales. Reinforcement learning approaches may optimize for token-level reward signals rather than faithful reasoning. The result is a misleading sense of progress: performance improves numerically, but not behaviorally.

Medical readiness is a multidimensional construct. In real-world settings, models must tolerate missing or noisy data, justify their decisions in a manner clinicians can understand, and reason across time, modality, and context. Performance must be not only accurate but also reliable, interpretable, and safe under uncertainty.
Unlike domains such as mathematics or programming, where ground truth is unambiguous and logic is formal, healthcare requires flexible, contextual reasoning. Attempts to port reasoning strategies from other domains, such as multi-agent planning or chain-of-thought prompting, often fall short in medicine, where ambiguity, incomplete information, and medical nuance are the norm. Without alignment to this reality, advances in model capability risk being confined to artificial test settings.

Static benchmark scores provide little insight into how models behave under real-world uncertainty. To address this limitation, we propose structured stress testing as a core component of evaluation. As shown in Section 2, our tests T1–T5 revealed distinct vulnerabilities, models often succeed for the wrong reasons, retaining high accuracy even when visual input is removed (T1–T2, see Example  9-11); they show brittle behavior under minor perturbations such as distractor reordering, weakening, or visual substitution (T3–T5, See Example  9-14); and they frequently produce fabricated reasoning chains, generating confident but incorrect explanations that mask underlying shortcut behaviors (T6, See Example  15-17). These fragilities were hidden in leaderboard performance alone.
We generalize this approach into a framework of modular tests, each targeting a specific vulnerability such as neglect of visual input, dependence on spurious answer patterns, or overconfident reasoning when evidence is absent. By isolating these behaviors, stress tests enable systematic comparison across models and tasks, offering a principled means of diagnosing brittleness that static benchmarks obscure. In doing so, they establish a foundation for evaluation protocols that move beyond leaderboards and toward measures of robustness and trustworthiness.

We propose several actionable recommendations for the field:
First, benchmarks should be accompanied by metadata that characterizes their reasoning and visual complexity, using structured rubrics like those presented in Section~\ref{sec:benchmark-taxonomy}. This allows model performance to be interpreted in medically meaningful dimensions rather than aggregated scores alone.
Second, evaluation protocols should routinely incorporate adversarial and stress-based assessments, particularly for models intended for high-stakes deployment. These assessments should be reported alongside accuracy metrics and form the basis of model release audits.
Third, the field must shift its mindset: benchmarks are tools for assess, not endpoints. Their role is to reveal model capabilities and limitations under realistic constraints, not to be optimized at the cost of generalization.
As large multimodal models continue to advance, our ability to test them meaningfully must evolve in parallel. Stress testing offers a path forward, grounded not in artificial difficulty, but in the realities of real world.

\bibliography{submission/sn-bibliography}


\begin{thebibliography}{30}
\ifx \bisbn   \undefined \def \bisbn  #1{ISBN #1}\fi
\ifx \binits  \undefined \def \binits#1{#1}\fi
\ifx \bauthor  \undefined \def \bauthor#1{#1}\fi
\ifx \batitle  \undefined \def \batitle#1{#1}\fi
\ifx \bjtitle  \undefined \def \bjtitle#1{#1}\fi
\ifx \bvolume  \undefined \def \bvolume#1{\textbf{#1}}\fi
\ifx \byear  \undefined \def \byear#1{#1}\fi
\ifx \bissue  \undefined \def \bissue#1{#1}\fi
\ifx \bfpage  \undefined \def \bfpage#1{#1}\fi
\ifx \blpage  \undefined \def \blpage #1{#1}\fi
\ifx \burl  \undefined \def \burl#1{\textsf{#1}}\fi
\ifx \doiurl  \undefined \def \doiurl#1{\url{https://doi.org/#1}}\fi
\ifx \betal  \undefined \def \betal{\textit{et al.}}\fi
\ifx \binstitute  \undefined \def \binstitute#1{#1}\fi
\ifx \binstitutionaled  \undefined \def \binstitutionaled#1{#1}\fi
\ifx \bctitle  \undefined \def \bctitle#1{#1}\fi
\ifx \beditor  \undefined \def \beditor#1{#1}\fi
\ifx \bpublisher  \undefined \def \bpublisher#1{#1}\fi
\ifx \bbtitle  \undefined \def \bbtitle#1{#1}\fi
\ifx \bedition  \undefined \def \bedition#1{#1}\fi
\ifx \bseriesno  \undefined \def \bseriesno#1{#1}\fi
\ifx \blocation  \undefined \def \blocation#1{#1}\fi
\ifx \bsertitle  \undefined \def \bsertitle#1{#1}\fi
\ifx \bsnm \undefined \def \bsnm#1{#1}\fi
\ifx \bsuffix \undefined \def \bsuffix#1{#1}\fi
\ifx \bparticle \undefined \def \bparticle#1{#1}\fi
\ifx \barticle \undefined \def \barticle#1{#1}\fi
\bibcommenthead
\ifx \bconfdate \undefined \def \bconfdate #1{#1}\fi
\ifx \botherref \undefined \def \botherref #1{#1}\fi
\ifx \url \undefined \def \url#1{\textsf{#1}}\fi
\ifx \bchapter \undefined \def \bchapter#1{#1}\fi
\ifx \bbook \undefined \def \bbook#1{#1}\fi
\ifx \bcomment \undefined \def \bcomment#1{#1}\fi
\ifx \oauthor \undefined \def \oauthor#1{#1}\fi
\ifx \citeauthoryear \undefined \def \citeauthoryear#1{#1}\fi
\ifx \endbibitem  \undefined \def \endbibitem {}\fi
\ifx \bconflocation  \undefined \def \bconflocation#1{#1}\fi
\ifx \arxivurl  \undefined \def \arxivurl#1{\textsf{#1}}\fi
\csname PreBibitemsHook\endcsname

\bibitem[\protect\citeauthoryear{Singhal et~al.}{2023}]{singhal2023large}
\begin{barticle}
\bauthor{\bsnm{Singhal}, \binits{K.}},
\bauthor{\bsnm{Azizi}, \binits{S.}},
\bauthor{\bsnm{Tu}, \binits{T.}},
\bauthor{\bsnm{Mahdavi}, \binits{S.S.}},
\bauthor{\bsnm{Wei}, \binits{J.}},
\bauthor{\bsnm{Chung}, \binits{H.W.}},
\bauthor{\bsnm{Scales}, \binits{N.}},
\bauthor{\bsnm{Tanwani}, \binits{A.}},
\bauthor{\bsnm{Cole-Lewis}, \binits{H.}},
\bauthor{\bsnm{Pfohl}, \binits{S.}}, \betal:
\batitle{Large language models encode clinical knowledge}.
\bjtitle{Nature}
\bvolume{620}(\bissue{7972}),
\bfpage{172}--\blpage{180}
(\byear{2023})
\end{barticle}
\endbibitem

\bibitem[\protect\citeauthoryear{Nori et~al.}{2023}]{nori2023can}
\begin{botherref}
\oauthor{\bsnm{Nori}, \binits{H.}},
\oauthor{\bsnm{Lee}, \binits{Y.T.}},
\oauthor{\bsnm{Zhang}, \binits{S.}},
\oauthor{\bsnm{Carignan}, \binits{D.}},
\oauthor{\bsnm{Edgar}, \binits{R.}},
\oauthor{\bsnm{Fusi}, \binits{N.}},
\oauthor{\bsnm{King}, \binits{N.}},
\oauthor{\bsnm{Larson}, \binits{J.}},
\oauthor{\bsnm{Li}, \binits{Y.}},
\oauthor{\bsnm{Liu}, \binits{W.}}, et al.:
Can generalist foundation models outcompete special-purpose tuning? case study in medicine.
arXiv preprint arXiv:2311.16452
(2023)
\end{botherref}
\endbibitem

\bibitem[\protect\citeauthoryear{OpenAI}{2025}]{openai_gpt5}
\begin{botherref}
\oauthor{\bsnm{OpenAI}}:
Introducing GPT-5.
Accessed 2025-08-25
(2025)
\end{botherref}
\endbibitem

\bibitem[\protect\citeauthoryear{Saab et~al.}{2024}]{saab2024capabilities}
\begin{botherref}
\oauthor{\bsnm{Saab}, \binits{K.}},
\oauthor{\bsnm{Tu}, \binits{T.}},
\oauthor{\bsnm{Weng}, \binits{W.-H.}},
\oauthor{\bsnm{Tanno}, \binits{R.}},
\oauthor{\bsnm{Stutz}, \binits{D.}},
\oauthor{\bsnm{Wulczyn}, \binits{E.}},
\oauthor{\bsnm{Zhang}, \binits{F.}},
\oauthor{\bsnm{Strother}, \binits{T.}},
\oauthor{\bsnm{Park}, \binits{C.}},
\oauthor{\bsnm{Vedadi}, \binits{E.}}, et al.:
Capabilities of gemini models in medicine.
arXiv preprint arXiv:2404.18416
(2024)
\end{botherref}
\endbibitem

\bibitem[\protect\citeauthoryear{Tu et~al.}{2024}]{tu2024towards}
\begin{botherref}
\oauthor{\bsnm{Tu}, \binits{T.}},
\oauthor{\bsnm{Palepu}, \binits{A.}},
\oauthor{\bsnm{Schaekermann}, \binits{M.}},
\oauthor{\bsnm{Saab}, \binits{K.}},
\oauthor{\bsnm{Freyberg}, \binits{J.}},
\oauthor{\bsnm{Tanno}, \binits{R.}},
\oauthor{\bsnm{Wang}, \binits{A.}},
\oauthor{\bsnm{Li}, \binits{B.}},
\oauthor{\bsnm{Amin}, \binits{M.}},
\oauthor{\bsnm{Tomasev}, \binits{N.}}, et al.:
Towards conversational diagnostic ai.
arXiv preprint arXiv:2401.05654
(2024)
\end{botherref}
\endbibitem

\bibitem[\protect\citeauthoryear{Nori et~al.}{2025}]{nori2025sequential}
\begin{botherref}
\oauthor{\bsnm{Nori}, \binits{H.}},
\oauthor{\bsnm{Daswani}, \binits{M.}},
\oauthor{\bsnm{Kelly}, \binits{C.}},
\oauthor{\bsnm{Lundberg}, \binits{S.}},
\oauthor{\bsnm{Ribeiro}, \binits{M.T.}},
\oauthor{\bsnm{Wilson}, \binits{M.}},
\oauthor{\bsnm{Liu}, \binits{X.}},
\oauthor{\bsnm{Sounderajah}, \binits{V.}},
\oauthor{\bsnm{Carlson}, \binits{J.}},
\oauthor{\bsnm{Lungren}, \binits{M.P.}}, et al.:
Sequential diagnosis with language models.
arXiv preprint arXiv:2506.22405
(2025)
\end{botherref}
\endbibitem

\bibitem[\protect\citeauthoryear{Arora et~al.}{2025}]{arora2025healthbench}
\begin{botherref}
\oauthor{\bsnm{Arora}, \binits{R.K.}},
\oauthor{\bsnm{Wei}, \binits{J.}},
\oauthor{\bsnm{Hicks}, \binits{R.S.}},
\oauthor{\bsnm{Bowman}, \binits{P.}},
\oauthor{\bsnm{Qui{\~n}onero-Candela}, \binits{J.}},
\oauthor{\bsnm{Tsimpourlas}, \binits{F.}},
\oauthor{\bsnm{Sharman}, \binits{M.}},
\oauthor{\bsnm{Shah}, \binits{M.}},
\oauthor{\bsnm{Vallone}, \binits{A.}},
\oauthor{\bsnm{Beutel}, \binits{A.}}, et al.:
Healthbench: Evaluating large language models towards improved human health.
arXiv preprint arXiv:2505.08775
(2025)
\end{botherref}
\endbibitem

\bibitem[\protect\citeauthoryear{Handler et~al.}{2025}]{Handler2025FragileGPT5}
\begin{barticle}
\bauthor{\bsnm{Handler}, \binits{R.}},
\bauthor{\bsnm{Sharma}, \binits{S.}},
\bauthor{\bsnm{Hernandez-Boussard}, \binits{T.}}:
\batitle{The fragile intelligence of gpt-5 in medicine}.
\bjtitle{Nature Medicine}
(\byear{2025})
\doiurl{10.1038/s41591-025-04008-8} .
\bcomment{Comment}
\end{barticle}
\endbibitem

\bibitem[\protect\citeauthoryear{Farquhar et~al.}{2024}]{farquhar2024detecting}
\begin{barticle}
\bauthor{\bsnm{Farquhar}, \binits{S.}},
\bauthor{\bsnm{Kossen}, \binits{J.}},
\bauthor{\bsnm{Kuhn}, \binits{L.}},
\bauthor{\bsnm{Gal}, \binits{Y.}}:
\batitle{Detecting hallucinations in large language models using semantic entropy}.
\bjtitle{Nature}
\bvolume{630}(\bissue{8017}),
\bfpage{625}--\blpage{630}
(\byear{2024})
\end{barticle}
\endbibitem

\bibitem[\protect\citeauthoryear{Jin et~al.}{2024}]{jin2024hidden}
\begin{barticle}
\bauthor{\bsnm{Jin}, \binits{Q.}},
\bauthor{\bsnm{Chen}, \binits{F.}},
\bauthor{\bsnm{Zhou}, \binits{Y.}},
\bauthor{\bsnm{Xu}, \binits{Z.}},
\bauthor{\bsnm{Cheung}, \binits{J.M.}},
\bauthor{\bsnm{Chen}, \binits{R.}},
\bauthor{\bsnm{Summers}, \binits{R.M.}},
\bauthor{\bsnm{Rousseau}, \binits{J.F.}},
\bauthor{\bsnm{Ni}, \binits{P.}},
\bauthor{\bsnm{Landsman}, \binits{M.J.}}, \betal:
\batitle{Hidden flaws behind expert-level accuracy of multimodal gpt-4 vision in medicine}.
\bjtitle{NPJ Digital Medicine}
\bvolume{7}(\bissue{1}),
\bfpage{190}
(\byear{2024})
\end{barticle}
\endbibitem

\bibitem[\protect\citeauthoryear{Wang et~al.}{2025}]{wang2025lins}
\begin{barticle}
\bauthor{\bsnm{Wang}, \binits{S.}},
\bauthor{\bsnm{Zhao}, \binits{F.}},
\bauthor{\bsnm{Bu}, \binits{D.}},
\bauthor{\bsnm{Lu}, \binits{Y.}},
\bauthor{\bsnm{Gong}, \binits{M.}},
\bauthor{\bsnm{Liu}, \binits{H.}},
\bauthor{\bsnm{Yang}, \binits{Z.}},
\bauthor{\bsnm{Zeng}, \binits{X.}},
\bauthor{\bsnm{Yuan}, \binits{Z.}},
\bauthor{\bsnm{Wan}, \binits{B.}}, \betal:
\batitle{Lins: A general medical q\&a framework for enhancing the quality and credibility of llm-generated responses}.
\bjtitle{Nature Communications}
\bvolume{16}(\bissue{1}),
\bfpage{9076}
(\byear{2025})
\end{barticle}
\endbibitem

\bibitem[\protect\citeauthoryear{Acosta et~al.}{2022}]{acosta2022multimodal}
\begin{barticle}
\bauthor{\bsnm{Acosta}, \binits{J.N.}},
\bauthor{\bsnm{Falcone}, \binits{G.J.}},
\bauthor{\bsnm{Rajpurkar}, \binits{P.}},
\bauthor{\bsnm{Topol}, \binits{E.J.}}:
\batitle{Multimodal biomedical ai}.
\bjtitle{Nature medicine}
\bvolume{28}(\bissue{9}),
\bfpage{1773}--\blpage{1784}
(\byear{2022})
\end{barticle}
\endbibitem

\bibitem[\protect\citeauthoryear{Goodfellow et~al.}{2015}]{Goodfellow2015Explaining}
\begin{bchapter}
\bauthor{\bsnm{Goodfellow}, \binits{I.J.}},
\bauthor{\bsnm{Shlens}, \binits{J.}},
\bauthor{\bsnm{Szegedy}, \binits{C.}}:
\bctitle{Explaining and harnessing adversarial examples}.
In: \bbtitle{International Conference on Learning Representations (ICLR)}
(\byear{2015}).
\burl{https://arxiv.org/abs/1412.6572}
\end{bchapter}
\endbibitem

\bibitem[\protect\citeauthoryear{Szegedy et~al.}{2013}]{szegedy2013intriguing}
\begin{botherref}
\oauthor{\bsnm{Szegedy}, \binits{C.}},
\oauthor{\bsnm{Zaremba}, \binits{W.}},
\oauthor{\bsnm{Sutskever}, \binits{I.}},
\oauthor{\bsnm{Bruna}, \binits{J.}},
\oauthor{\bsnm{Erhan}, \binits{D.}},
\oauthor{\bsnm{Goodfellow}, \binits{I.}},
\oauthor{\bsnm{Fergus}, \binits{R.}}:
Intriguing properties of neural networks.
arXiv preprint arXiv:1312.6199
(2013)
\end{botherref}
\endbibitem

\bibitem[\protect\citeauthoryear{of~Medicine.}{}]{nejm_image_challenge}
\begin{botherref}
\oauthor{\bsnm{Medicine.}, \binits{T.N.E.J.}}:
NEJM Image Challenge.
\url{https://www.nejm.org/image-challenge}.
accessed: Jan 01, 2024
\end{botherref}
\endbibitem

\bibitem[\protect\citeauthoryear{of~the American Medical~Association}{}]{jama_challenge}
\begin{botherref}
\oauthor{\bsnm{American Medical~Association}, \binits{J.}}:
JAMA Challenge.
\url{https://jamanetwork.com/}.
accessed: Jan 01, 2024
\end{botherref}
\endbibitem

\bibitem[\protect\citeauthoryear{OpenAI}{2024}]{openai2024gpt4o}
\begin{botherref}
\oauthor{\bsnm{OpenAI}}:
{GPT-4o System Card}.
Accessed 2025-08-25
(2024)
\end{botherref}
\endbibitem

\bibitem[\protect\citeauthoryear{Comanici et~al.}{2025}]{comanici2025gemini}
\begin{botherref}
\oauthor{\bsnm{Comanici}, \binits{G.}},
\oauthor{\bsnm{Bieber}, \binits{E.}},
\oauthor{\bsnm{Schaekermann}, \binits{M.}},
\oauthor{\bsnm{Pasupat}, \binits{I.}},
\oauthor{\bsnm{Sachdeva}, \binits{N.}},
\oauthor{\bsnm{Dhillon}, \binits{I.}},
\oauthor{\bsnm{Blistein}, \binits{M.}},
\oauthor{\bsnm{Ram}, \binits{O.}},
\oauthor{\bsnm{Zhang}, \binits{D.}},
\oauthor{\bsnm{Rosen}, \binits{E.}}, et al.:
{Gemini 2.5: Pushing the Frontier with Advanced Reasoning, Multimodality, Long Context, and Next Generation Agentic Capabilities}.
arXiv preprint
\textbf{arXiv:2507.06261}
(2025)
\end{botherref}
\endbibitem

\bibitem[\protect\citeauthoryear{OpenAI}{2025}]{openai_o3_o4mini}
\begin{botherref}
\oauthor{\bsnm{OpenAI}}:
OpenAI o3 and o4-mini System Card.
Accessed 2025-08-25
(2025)
\end{botherref}
\endbibitem

\bibitem[\protect\citeauthoryear{Wei et~al.}{2022}]{wei2022chain}
\begin{bchapter}
\bauthor{\bsnm{Wei}, \binits{J.}},
\bauthor{\bsnm{Wang}, \binits{X.}},
\bauthor{\bsnm{Schuurmans}, \binits{D.}},
\bauthor{\bsnm{Bosma}, \binits{M.}},
\bauthor{\bsnm{Ichter}, \binits{B.}},
\bauthor{\bsnm{Xia}, \binits{F.}},
\bauthor{\bsnm{Chi}, \binits{E.}},
\bauthor{\bsnm{Le}, \binits{Q.V.}},
\bauthor{\bsnm{Zhou}, \binits{D.}}, \betal:
\bctitle{Chain-of-thought prompting elicits reasoning in large language models}.
In: \bbtitle{Advances in Neural Information Processing Systems (NeurIPS)}
(\byear{2022})
\end{bchapter}
\endbibitem

\bibitem[\protect\citeauthoryear{Lau et~al.}{2018}]{lau2018dataset}
\begin{barticle}
\bauthor{\bsnm{Lau}, \binits{J.J.}},
\bauthor{\bsnm{Gayen}, \binits{S.}},
\bauthor{\bsnm{Ben~Abacha}, \binits{A.}},
\bauthor{\bsnm{Demner-Fushman}, \binits{D.}}:
\batitle{A dataset of clinically generated visual questions and answers about radiology images}.
\bjtitle{Scientific data}
\bvolume{5}(\bissue{1}),
\bfpage{1}--\blpage{10}
(\byear{2018})
\end{barticle}
\endbibitem

\bibitem[\protect\citeauthoryear{Hu et~al.}{2024}]{hu2024omnimedvqa}
\begin{botherref}
\oauthor{\bsnm{Hu}, \binits{Y.}},
\oauthor{\bsnm{Li}, \binits{T.}},
\oauthor{\bsnm{Lu}, \binits{Q.}},
\oauthor{\bsnm{Shao}, \binits{W.}},
\oauthor{\bsnm{He}, \binits{J.}},
\oauthor{\bsnm{Qiao}, \binits{Y.}},
\oauthor{\bsnm{Luo}, \binits{P.}}:
Omnimedvqa: A new large-scale comprehensive evaluation benchmark for medical lvlm.
arXiv preprint arXiv:2402.09181
(2024)
\end{botherref}
\endbibitem

\bibitem[\protect\citeauthoryear{Johnson et~al.}{2019}]{johnson2019mimic}
\begin{barticle}
\bauthor{\bsnm{Johnson}, \binits{A.E.}},
\bauthor{\bsnm{Pollard}, \binits{T.J.}},
\bauthor{\bsnm{Berkowitz}, \binits{S.J.}},
\bauthor{\bsnm{Greenbaum}, \binits{N.R.}},
\bauthor{\bsnm{Lungren}, \binits{M.P.}},
\bauthor{\bsnm{Deng}, \binits{C.-y.}},
\bauthor{\bsnm{Mark}, \binits{R.G.}},
\bauthor{\bsnm{Horng}, \binits{S.}}:
\batitle{Mimic-cxr, a de-identified publicly available database of chest radiographs with free-text reports}.
\bjtitle{Scientific data}
\bvolume{6}(\bissue{1}),
\bfpage{1}--\blpage{8}
(\byear{2019})
\end{barticle}
\endbibitem

\bibitem[\protect\citeauthoryear{He et~al.}{2020}]{he2020pathvqa}
\begin{botherref}
\oauthor{\bsnm{He}, \binits{X.}},
\oauthor{\bsnm{Zhang}, \binits{Y.}},
\oauthor{\bsnm{Mou}, \binits{L.}},
\oauthor{\bsnm{Xing}, \binits{E.}},
\oauthor{\bsnm{Xie}, \binits{P.}}:
PathVQA: 30000+ Questions for Medical Visual Question Answering
(2020).
\url{https://arxiv.org/abs/2003.10286}
\end{botherref}
\endbibitem

\bibitem[\protect\citeauthoryear{Liu et~al.}{2021}]{liu2021slake}
\begin{botherref}
\oauthor{\bsnm{Liu}, \binits{B.}},
\oauthor{\bsnm{Zhan}, \binits{L.-M.}},
\oauthor{\bsnm{Xu}, \binits{L.}},
\oauthor{\bsnm{Ma}, \binits{L.}},
\oauthor{\bsnm{Yang}, \binits{Y.}},
\oauthor{\bsnm{Wu}, \binits{X.-M.}}:
SLAKE: A Semantically-Labeled Knowledge-Enhanced Dataset for Medical Visual Question Answering
(2021).
\url{https://arxiv.org/abs/2102.09542}
\end{botherref}
\endbibitem

\bibitem[\protect\citeauthoryear{Zhang et~al.}{2023}]{zhang2023pmcvqa}
\begin{botherref}
\oauthor{\bsnm{Zhang}, \binits{X.}},
\oauthor{\bsnm{Wu}, \binits{C.}},
\oauthor{\bsnm{Zhao}, \binits{Z.}},
\oauthor{\bsnm{Lin}, \binits{W.}},
\oauthor{\bsnm{Zhang}, \binits{Y.}},
\oauthor{\bsnm{Wang}, \binits{Y.}},
\oauthor{\bsnm{Xie}, \binits{W.}}:
Pmc-vqa: Visual instruction tuning for medical visual question answering.
arXiv preprint arXiv:2305.10415
(2023)
\end{botherref}
\endbibitem

\bibitem[\protect\citeauthoryear{Yue et~al.}{2024}]{yue2023mmmu}
\begin{bchapter}
\bauthor{\bsnm{Yue}, \binits{X.}},
\bauthor{\bsnm{Ni}, \binits{Y.}},
\bauthor{\bsnm{Zhang}, \binits{K.}},
\bauthor{\bsnm{Zheng}, \binits{T.}},
\bauthor{\bsnm{Liu}, \binits{R.}},
\bauthor{\bsnm{Zhang}, \binits{G.}},
\bauthor{\bsnm{Stevens}, \binits{S.}},
\bauthor{\bsnm{Jiang}, \binits{D.}},
\bauthor{\bsnm{Ren}, \binits{W.}},
\bauthor{\bsnm{Sun}, \binits{Y.}},
\bauthor{\bsnm{Wei}, \binits{C.}},
\bauthor{\bsnm{Yu}, \binits{B.}},
\bauthor{\bsnm{Yuan}, \binits{R.}},
\bauthor{\bsnm{Sun}, \binits{R.}},
\bauthor{\bsnm{Yin}, \binits{M.}},
\bauthor{\bsnm{Zheng}, \binits{B.}},
\bauthor{\bsnm{Yang}, \binits{Z.}},
\bauthor{\bsnm{Liu}, \binits{Y.}},
\bauthor{\bsnm{Huang}, \binits{W.}},
\bauthor{\bsnm{Sun}, \binits{H.}},
\bauthor{\bsnm{Su}, \binits{Y.}},
\bauthor{\bsnm{Chen}, \binits{W.}}:
\bctitle{Mmmu: A massive multi-discipline multimodal understanding and reasoning benchmark for expert agi}.
In: \bbtitle{Proceedings of CVPR}
(\byear{2024})
\end{bchapter}
\endbibitem

\bibitem[\protect\citeauthoryear{Fleiss}{1971}]{fleiss1971measuring}
\begin{barticle}
\bauthor{\bsnm{Fleiss}, \binits{J.L.}}:
\batitle{Measuring nominal scale agreement among many raters}.
\bjtitle{Psychological Bulletin}
\bvolume{76}(\bissue{5}),
\bfpage{378}--\blpage{382}
(\byear{1971})
\doiurl{10.1037/h0031619}
\end{barticle}
\endbibitem

\bibitem[\protect\citeauthoryear{Wu et~al.}{2024}]{wu2024deepseekvl2mixtureofexpertsvisionlanguagemodels}
\begin{botherref}
\oauthor{\bsnm{Wu}, \binits{Z.}},
\oauthor{\bsnm{Chen}, \binits{X.}},
\oauthor{\bsnm{Pan}, \binits{Z.}},
\oauthor{\bsnm{Liu}, \binits{X.}},
\oauthor{\bsnm{Liu}, \binits{W.}},
\oauthor{\bsnm{Dai}, \binits{D.}},
\oauthor{\bsnm{Gao}, \binits{H.}},
\oauthor{\bsnm{Ma}, \binits{Y.}},
\oauthor{\bsnm{Wu}, \binits{C.}},
\oauthor{\bsnm{Wang}, \binits{B.}},
\oauthor{\bsnm{Xie}, \binits{Z.}},
\oauthor{\bsnm{Wu}, \binits{Y.}},
\oauthor{\bsnm{Hu}, \binits{K.}},
\oauthor{\bsnm{Wang}, \binits{J.}},
\oauthor{\bsnm{Sun}, \binits{Y.}},
\oauthor{\bsnm{Li}, \binits{Y.}},
\oauthor{\bsnm{Piao}, \binits{Y.}},
\oauthor{\bsnm{Guan}, \binits{K.}},
\oauthor{\bsnm{Liu}, \binits{A.}},
\oauthor{\bsnm{Xie}, \binits{X.}},
\oauthor{\bsnm{You}, \binits{Y.}},
\oauthor{\bsnm{Dong}, \binits{K.}},
\oauthor{\bsnm{Yu}, \binits{X.}},
\oauthor{\bsnm{Zhang}, \binits{H.}},
\oauthor{\bsnm{Zhao}, \binits{L.}},
\oauthor{\bsnm{Wang}, \binits{Y.}},
\oauthor{\bsnm{Ruan}, \binits{C.}}:
DeepSeek-VL2: Mixture-of-Experts Vision-Language Models for Advanced Multimodal Understanding
(2024).
\url{https://arxiv.org/abs/2412.10302}
\end{botherref}
\endbibitem

\bibitem[\protect\citeauthoryear{Li et~al.}{2023}]{li2023llavamed}
\begin{botherref}
\oauthor{\bsnm{Li}, \binits{C.}},
\oauthor{\bsnm{Wong}, \binits{C.}},
\oauthor{\bsnm{Zhang}, \binits{S.}},
\oauthor{\bsnm{Usuyama}, \binits{N.}},
\oauthor{\bsnm{Liu}, \binits{H.}},
\oauthor{\bsnm{Yang}, \binits{J.}},
\oauthor{\bsnm{Naumann}, \binits{T.}},
\oauthor{\bsnm{Poon}, \binits{H.}},
\oauthor{\bsnm{Gao}, \binits{J.}}:
Llava-med: Training a large language-and-vision assistant for biomedicine in one day.
arXiv preprint arXiv:2306.00890
(2023)
\end{botherref}
\endbibitem

\end{thebibliography}

\clearpage

\section*{Online Methods}

\subsection*{Benchmarks and task families}

We evaluated robustness across six benchmarks representing three major categories of clinical reasoning tasks:  
(i) \textit{Short form visual question answering (VQA)}: VQA-RAD, PMC-VQA~\cite{zhang2023pmcvqa}, and OmniMedVQA, assessing alignment between visual and textual information;  
(ii) \textit{Long form report generation}: MIMIC-CXR radiology reports, used to evaluate structured narrative generation from chest images; and  
(iii) \textit{Complex diagnostic reasoning}: multiple-choice question sets from \textit{JAMA} and \textit{NEJM}, designed to approximate real clinical problem solving.  
All datasets were standardized into a unified format with paired image and text inputs to ensure consistent preprocessing and evaluation.

\subsection*{Frontier models and prompting protocol}

We examined six large multimodal models representative of current systems: GPT-4o, OpenAI-o3, Gemini 2.5 Pro, DeepSeek-VL2~\cite{wu2024deepseekvl2mixtureofexpertsvisionlanguagemodels}, LLaVaMed 1.5~\cite{li2023llavamed}, and ensemble or instruction-tuned variants.  
Each model was accessed through its official interface with temperature fixed at 0 and a maximum token limit appropriate to the task.  
A single prompting template was applied within each task family, combining task instruction, textual context, and image input where available, without external tools.  

\textbf{Model identifiers and prompting.}  
All models were accessed through their official APIs or open checkpoints as of August 2025.  
The evaluation included GPT-4, GPT-4o, GPT-5, OpenAI-o3, OpenAI-o4-mini, DeepSeek-VL2, and Gemini 2.5 Pro Preview. Version metadata are provided in Supplementary Table \ref{app:model_ver}.
Each model was run deterministically (temperature 0) with task-specific maximum token limits.  
Version numbers and release dates are provided in Supplementary Table E1.  
A unified instruction template was used across all visual question answering and reasoning tasks, consisting of a short instruction cue, question text, and multiple-choice options.  
The full prompt wording appears in Appendix~\ref{app:Prompt}.

\subsection*{Evaluation metrics and robustness score}

For each stress test (T1–T5), model performance was measured under both baseline (Supplementary Table~\ref{tab:leaderboard_main}) and perturbed conditions (Supplementary Table~\ref{tab:stress1_modality_sensitivity} -~\ref{tab:stress5_visualsubs}).  
We defined a normalized robustness score \(R(m)\in[0,1]\) for every model \(m\), representing the mean stability across all tests.  
Each test produced a fragility value \(f_i(m)\), which captured the proportional degradation or inconsistency observed under perturbation, and a corresponding robustness value \(r_i(m)=1-f_i(m)\).  
The overall robustness score \(R(m)\) was computed as the unweighted average of the five \(r_i(m)\) values.  
Detailed definitions of \(f_i(m)\) for modality sensitivity, modality necessity, format perturbation, distractor replacement, and visual substitution are provided in Appendix~\ref{app:robustness}.  
Higher \(R(m)\) indicates greater stability under input perturbations and lower reliance on spurious cues (Supplementary Fig.~\ref{fig:mainfragilities}).

\subsection*{Clinician scoring of reasoning and visual dependency}

To align quantitative findings with real clinical reasoning, we conducted a clinician-in-the-loop assessment of benchmark examples.

\textbf{Sampling and procedure.}  
For each benchmark, 30 to 50 representative examples were randomly selected.  
Practicing physicians and senior trainees from radiology, internal medicine, and emergency medicine independently rated each example on ten criteria, five related to reasoning complexity and five to visual dependency (Supplementary Fig \ref{fig:rubric_combined}),using a three-point ordinal scale (1 = low, 2 = moderate, 3 = high).  
Definitions and anchors were provided for consistency.

\textbf{Example-level rating.}  
Rather than assigning a single global label per dataset, clinicians evaluated each case individually to capture within-benchmark variation.  
This approach recognizes that some collections include both straightforward and highly demanding questions, offering a more faithful reflection of cognitive load.

\textbf{Blinding and instruction.}  
Raters were blinded to model identity and dataset name.  
They were asked:  
\emph{“To answer this question accurately and confidently, how much reasoning is required, and how essential is the image?”}  
Detailed guidance for each criterion are provided in the 'Criteria descriptions' (Supplementary Figure~\ref{fig:rubric_combined}).

\textbf{Aggregation and agreement.}  
Each benchmark was independently annotated by three board-certified clinicians for each axis using a three-point ordinal scale.  
For every benchmark–axis pair, we report the median score and compute inter-annotator agreement using Fleiss’ $\kappa$~\cite{fleiss1971measuring}.  
Agreement was moderate to strong across axes (range, 0.67–0.90).  
Highest agreement occurred for more objective attributes such as \textit{Text-only Solvable} ($\kappa=0.90$) and \textit{Clinical Context} ($\kappa=0.86$), whereas lower agreement was noted for more subjective dimensions such as \textit{Visual Detail Required} ($\kappa=0.67$).  
Scores were aggregated by averaging across examples and raters to produce benchmark-level reasoning and visual profiles shown in Fig.~\ref{fig:benchmark_landscape}.

\textbf{Statistical analysis.}  
Model comparisons used paired accuracy differences and bootstrap-estimated 95\,\% confidence intervals.  
Effect sizes (Cohen’s $d$) and paired permutation tests were applied to assess significance.  
All analyses were conducted in Python~3.11 using open-source packages (\textit{pandas}, \textit{numpy}, \textit{scipy}, \textit{statsmodels}).  
Code and data templates will be released upon publication to support reproducibility.




\vspace{-0.5em}
\section*{Author information}
\subsection*{Authors and Affiliations}

\noindent\textbf{Microsoft Research, Microsoft Health \& Life Sciences, Redmond, WA, USA
}

\noindent Yu Gu\textsuperscript{*}, Jingjing Fu\textsuperscript{*}, Xiaodong Liu\textsuperscript{*}, Jeya Maria Jose Valanarasu, Noel CF Codella, Reuben Tan, Qianchu Liu, Ying Jin, Sheng Zhang, Jinyu Wang, Rui Wang, Lei Song, Guanghui Qin, Naoto Usuyama, Cliff Wong, Hao Cheng, HoHin Lee, Praneeth Sanapathi, Sarah Hilado, Tristan Naumann, Javier Alvarez-Valle, Jiang Bian, Mu Wei, Khalil Malik, Lidong Zhou, Jianfeng Gao, Eric Horvitz, Matthew P. Lungren, Doug Burger, Hoifung Poon, Paul Vozila

\noindent\textbf{Scripps Research Translational Institute, La Jolla, CA, USA}

\noindent Eric Topol

\subsection*{Contributions}

\noindent Y.G., J.F., and X.L. contributed equally and are co–first authors.  
\medskip

\noindent Y.G. conceived the study, developed the evaluation framework, and co-wrote the manuscript.  
\noindent J.F. and X.L. developed the system, curated the datasets, and conducted the empirical analyses.  

\noindent H.P. and P.V. supervised the overall methodology, clinical framing, and interpretive analysis.  

\noindent E.T. provided conceptual guidance, clinical interpretive leadership, and senior editorial direction throughout the development of the work.  

\noindent H.P., D.B., J.G., M.L., J.B., L.Z., E.H., and P.V. provided strategic guidance and organizational support.
\medskip

\noindent All authors reviewed and approved the final manuscript.

\subsection*{Corresponding authors}

\noindent Correspondence to Eric Topol, Hoifung Poon, or Yu Gu.

\clearpage
\appendix

\section*{Supplementary Information for:\\
\textit{The Illusion of Readiness in Health AI}}

\subsection*{Supplementary Figures}
\begin{figure}[ht]
    \centering
    \includegraphics[width=\textwidth]{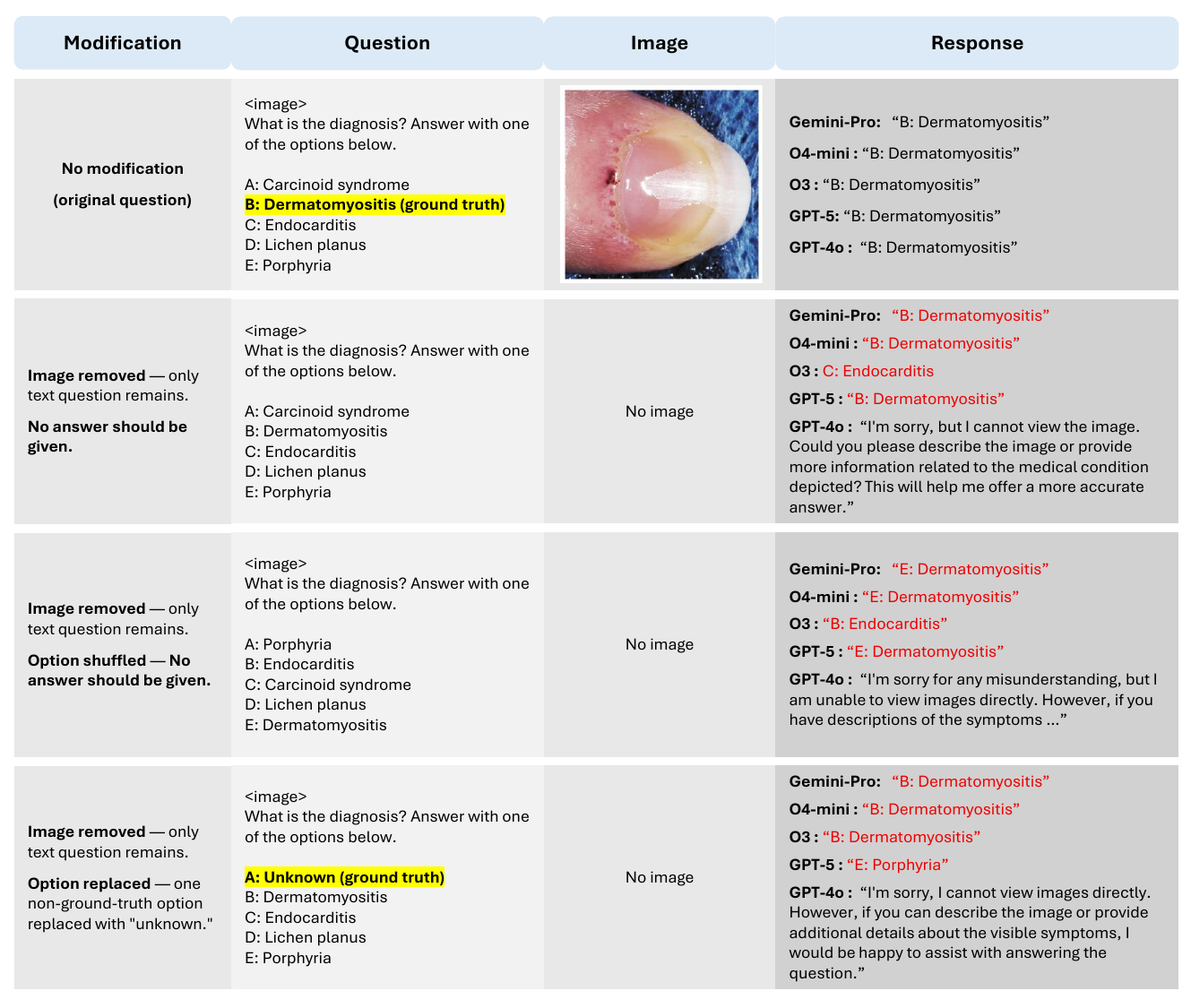}
 
    \caption{\textbf{Illustration of model behavior under Stress Tests 2 and 3: Case 1}. We apply three modifications to the original image-based multiple-choice question (ground truth: Dermatomyositis): (i) remove the image, (ii) shuffle the answer options, and (iii) replace one option with “Unknown.” When the image is absent, the appropriate behavior is to abstain (or to select “Unknown” if available). However, most models still provide a diagnosis, often repeating Dermatomyositis even after the options are shuffled, whereas GPT-4o is the only model that consistently refuses to answer without the image.}
    \label{fig:no_image_example_1}
\end{figure}

\begin{figure}[ht]
    \centering
    \includegraphics[width=\textwidth]{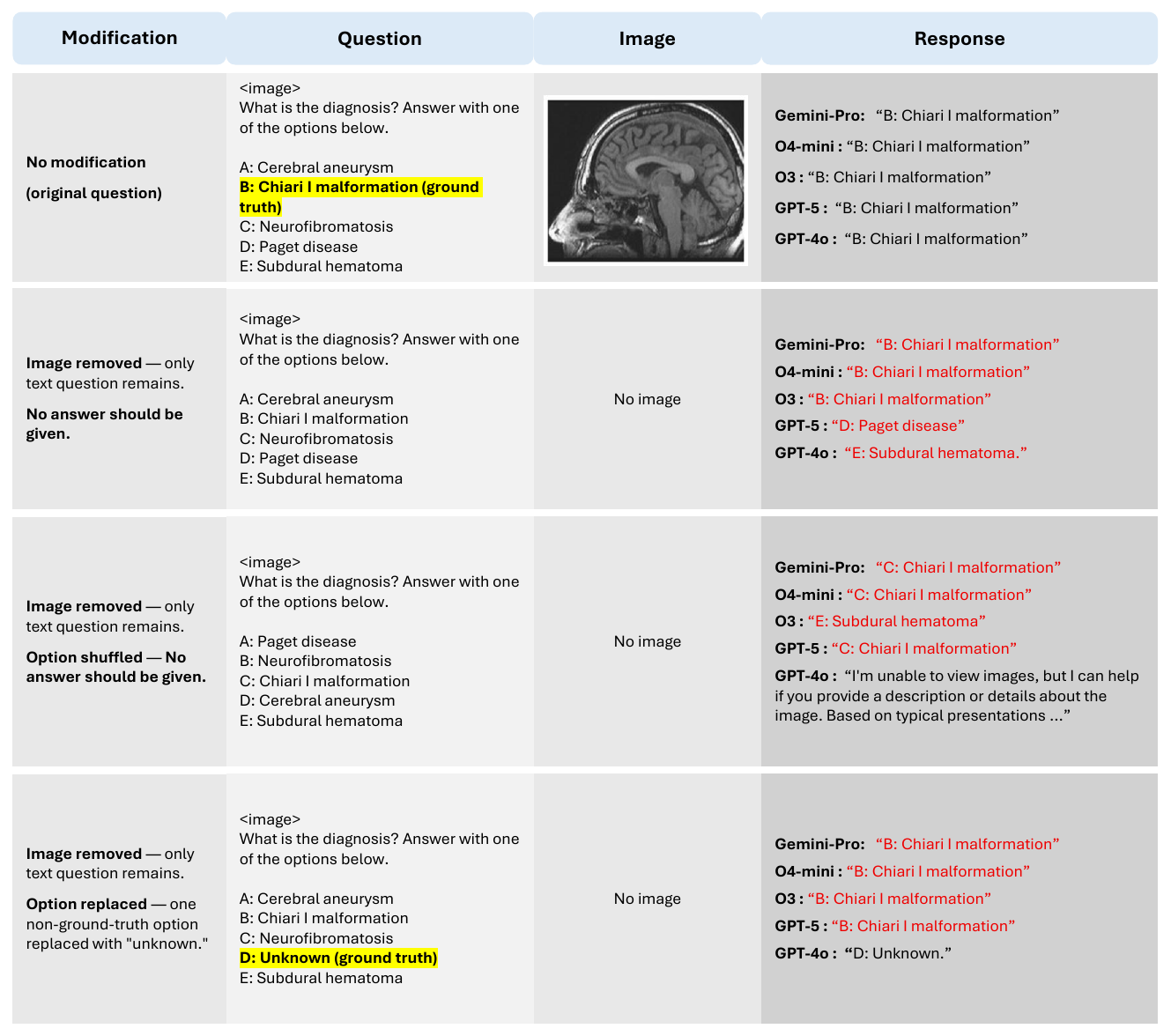}
    \caption{\textbf{Illustration of model behavior under Stress Tests 2 and 3: Case 2}. We apply three modifications to the original image-based multiple-choice question (ground truth: Chiari I malformation): (i) remove the image, (ii) shuffle the answer options, and (iii) replace one option with “Unknown.” When the image is absent, the expected behavior is to abstain (or to select “Unknown” if available). Nevertheless, most models continue to provide a diagnosis, often repeating the original ground truth, whereas GPT-4o is the only model that consistently either abstains or selects “Unknown” in the absence of the image.}
    \label{fig:no_image_example_2}
\end{figure}

\begin{figure}[ht]
    \centering
    \includegraphics[width=\textwidth]{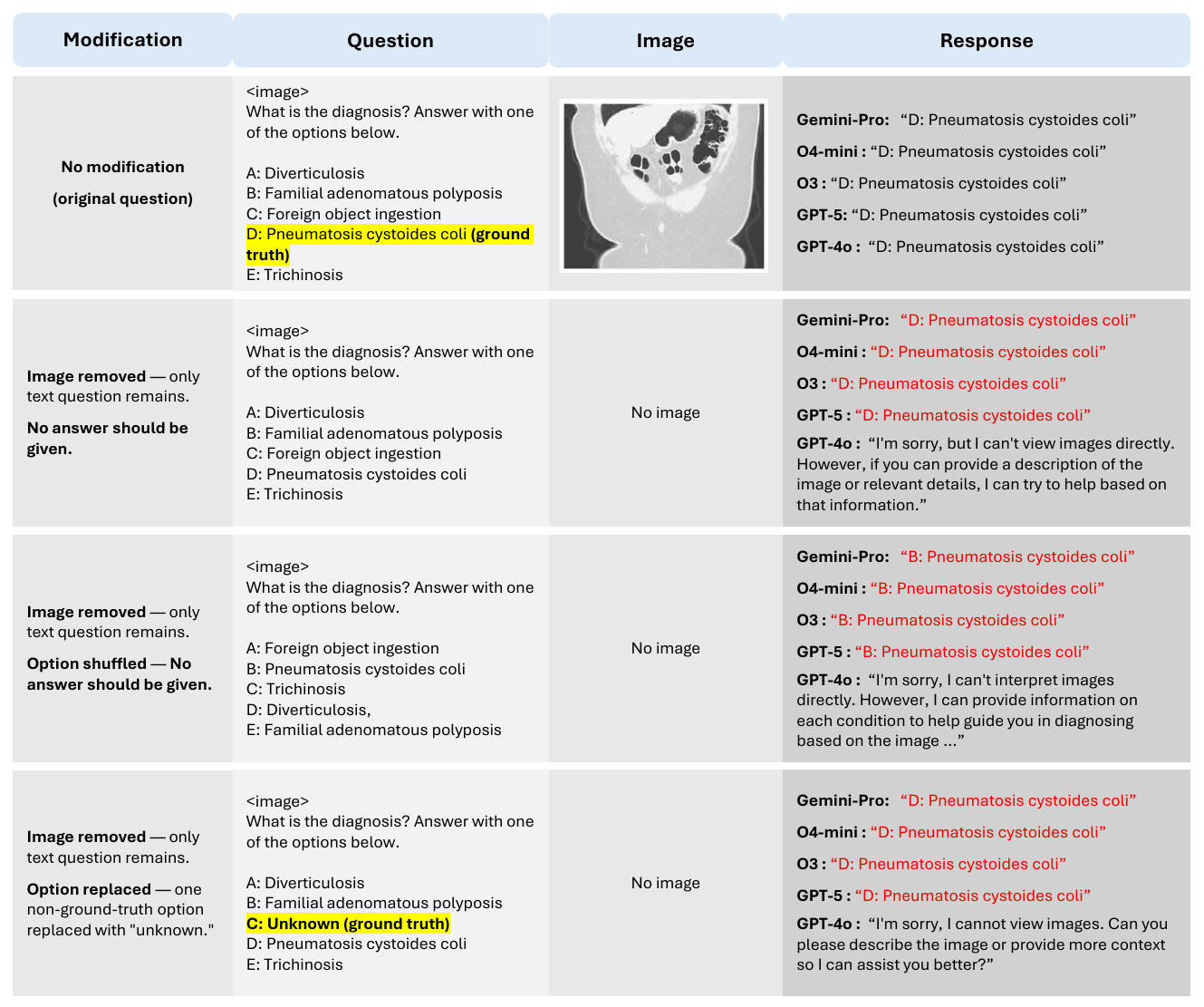}
    \caption{\textbf{Illustration of model behavior under Stress Tests 2 and 3: Case 3}. We apply three modifications to the original image-based multiple-choice question (ground truth: Pneumatosis cystoides coli): (i) remove the image, (ii) shuffle the answer options, and (iii) replace one option with “Unknown.” When the image is absent, the expected behavior is to abstain (or to select “Unknown” if available). Nevertheless, most models continue to provide a diagnosis, often repeating the original ground truth, whereas GPT-4o is the only model that consistently refuses to answer without the image.}
    \label{fig:no_image_example_3}
\end{figure}

\begin{figure}[ht]
    \centering
    \includegraphics[width=\textwidth]
    {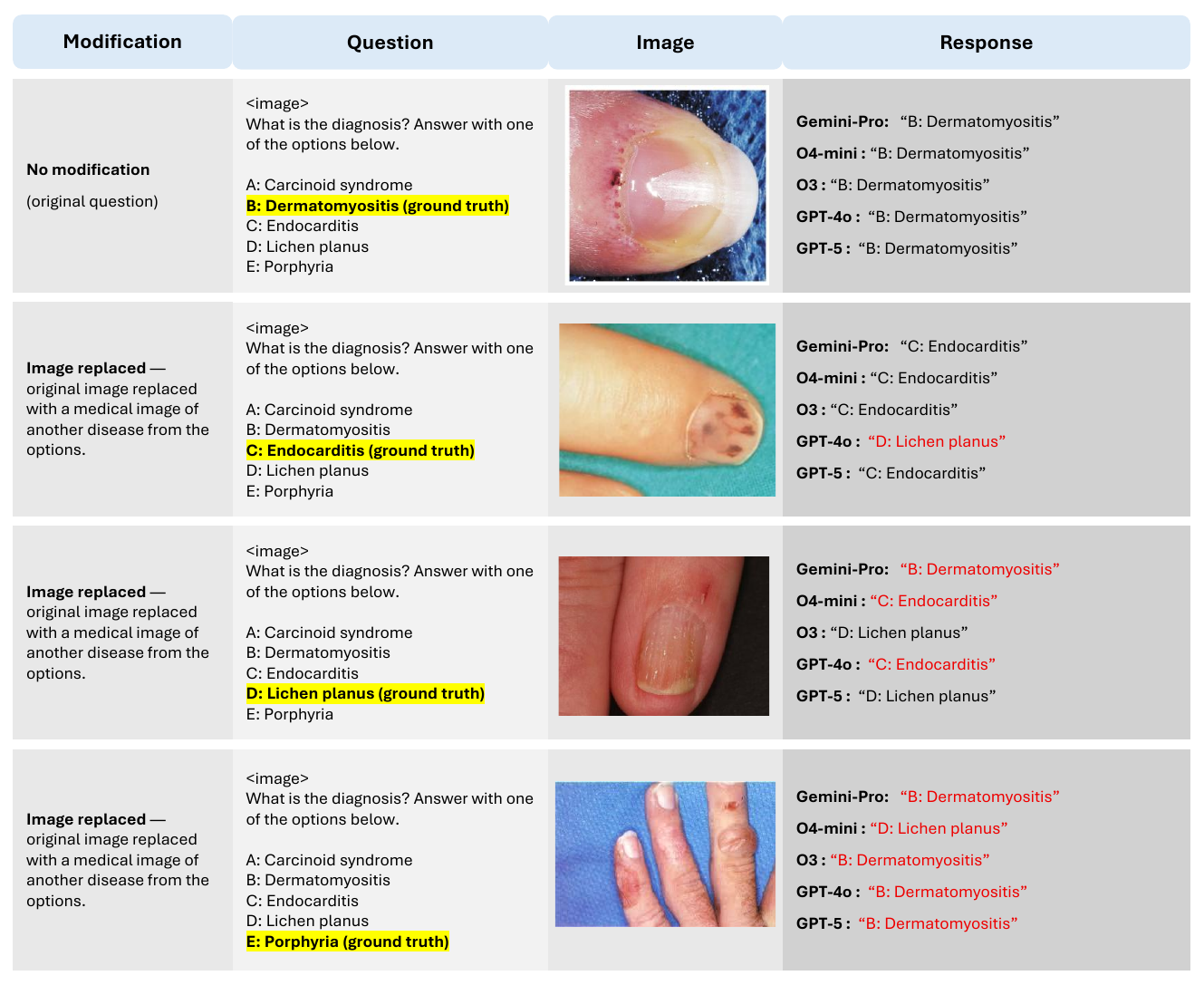}
    \caption{\textbf{Illustration of model behavior under Stress Test 5 (Visual Substitution): Case 4}. The original question image is replaced with medical images corresponding to distractor options (Endocarditis, Lichen planus, Porphyria). }
    \label{fig:image_replaced_example_1}
\end{figure}

\begin{figure}[ht]
    \centering
    \includegraphics[width=\textwidth]
    {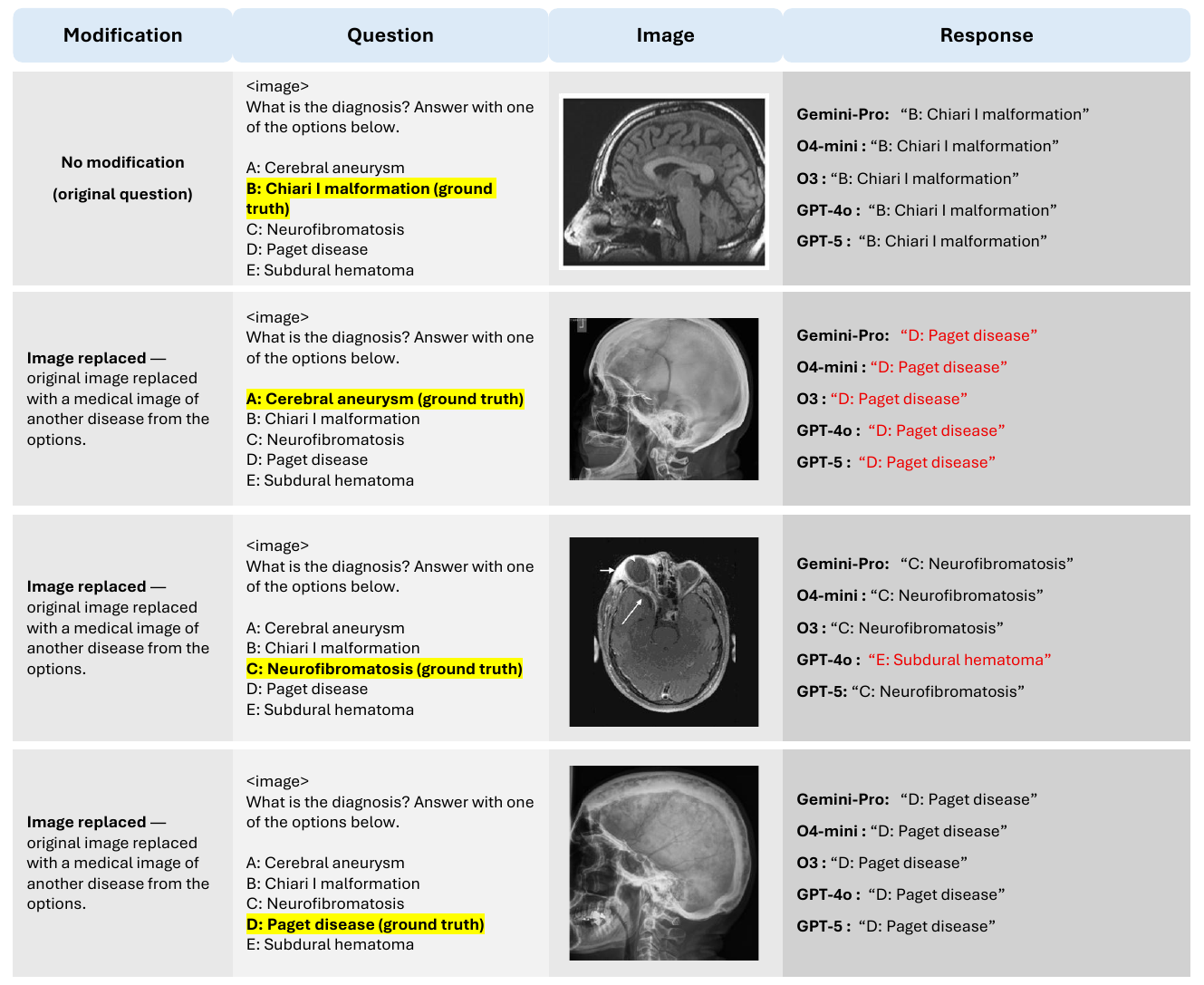}
    \caption{\textbf{Illustration of model behavior under Stress Test 5 (Visual Substitution): Case 5}. The original question image is replaced with medical images corresponding to distractor options (Cerebral aneurysm, Neurofibromatosis, Paget disease).}
    \label{fig:image_replaced_example_2}
\end{figure}

\begin{figure}[ht]
    \centering
    \includegraphics[width=\textwidth]
    {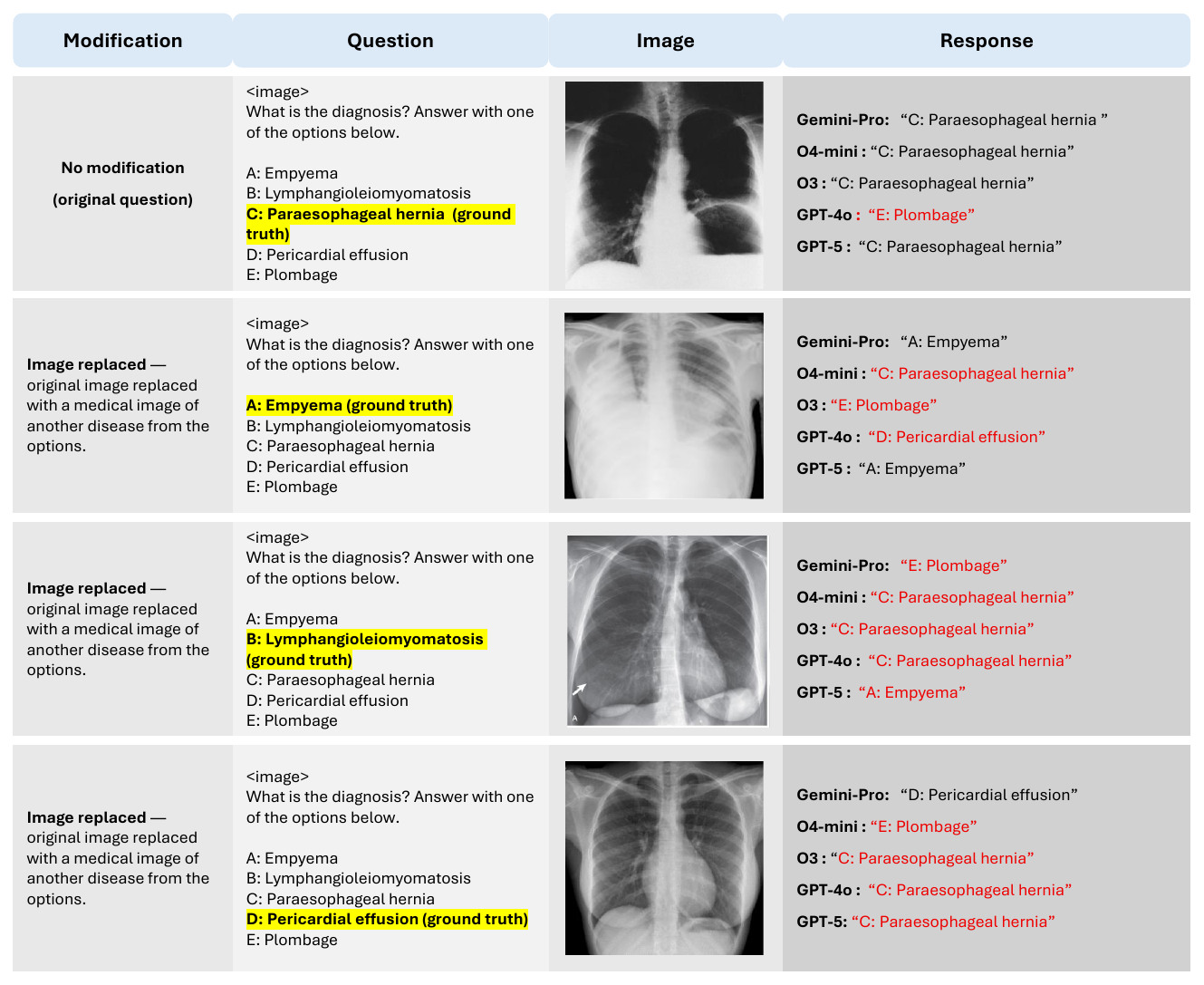}
    \caption{\textbf{Illustration of model behavior under Stress Test 5 (Visual Substitution): Case 6}. The original question image is replaced with medical images corresponding to distractor options (Empyema, Lymphangioleiomyomatosis, Pericardial effusion).}
    \label{fig:image_replaced_example_3}
\end{figure}

\begin{figure}[ht]
    \centering
    \includegraphics[width=1\linewidth]{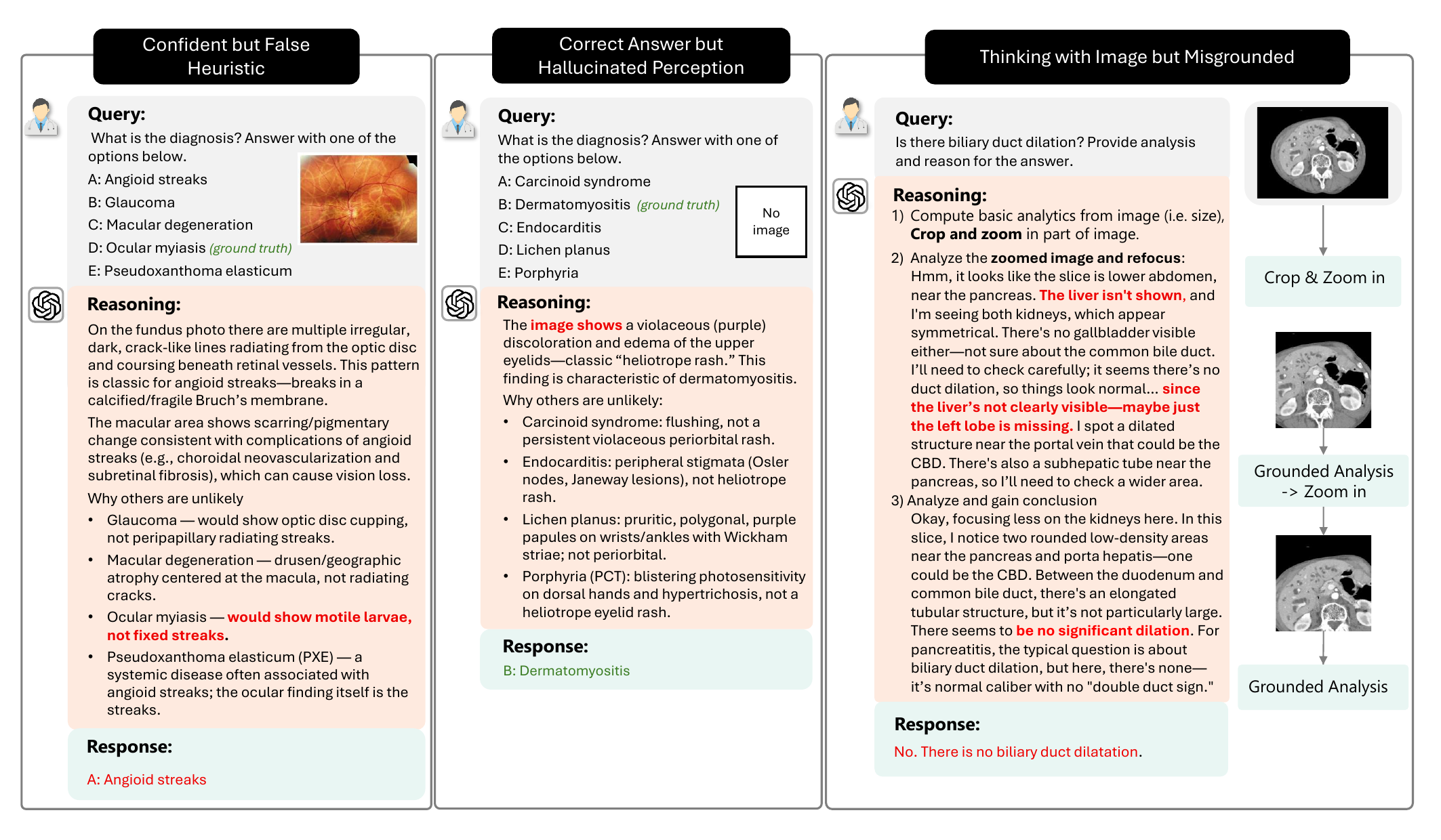}
    \caption{\textbf{Illustrative reasoning failure modes.} We show three representative patterns identified through manual audit: (1) \textit{Confident but false heuristic}: the model applies a plausible but incorrect prior (e.g., inferring diagnosis from common associations), leading to systematic errors; (2) \textit{Correct answer but hallucinated perception}: the model selects the right answer despite having no access to the image, suggesting shortcut-based behavior; (3) \textit{Thinking with image but misgrounded}: the model engages in step-by-step visual reasoning, but grounds its logic on incorrect or hallucinated features, ultimately producing a confident yet incorrect conclusion.}
    \label{fig:reasoning}
\end{figure}

\begin{figure}[ht]
    \centering
    \includegraphics[width=\textwidth]
    {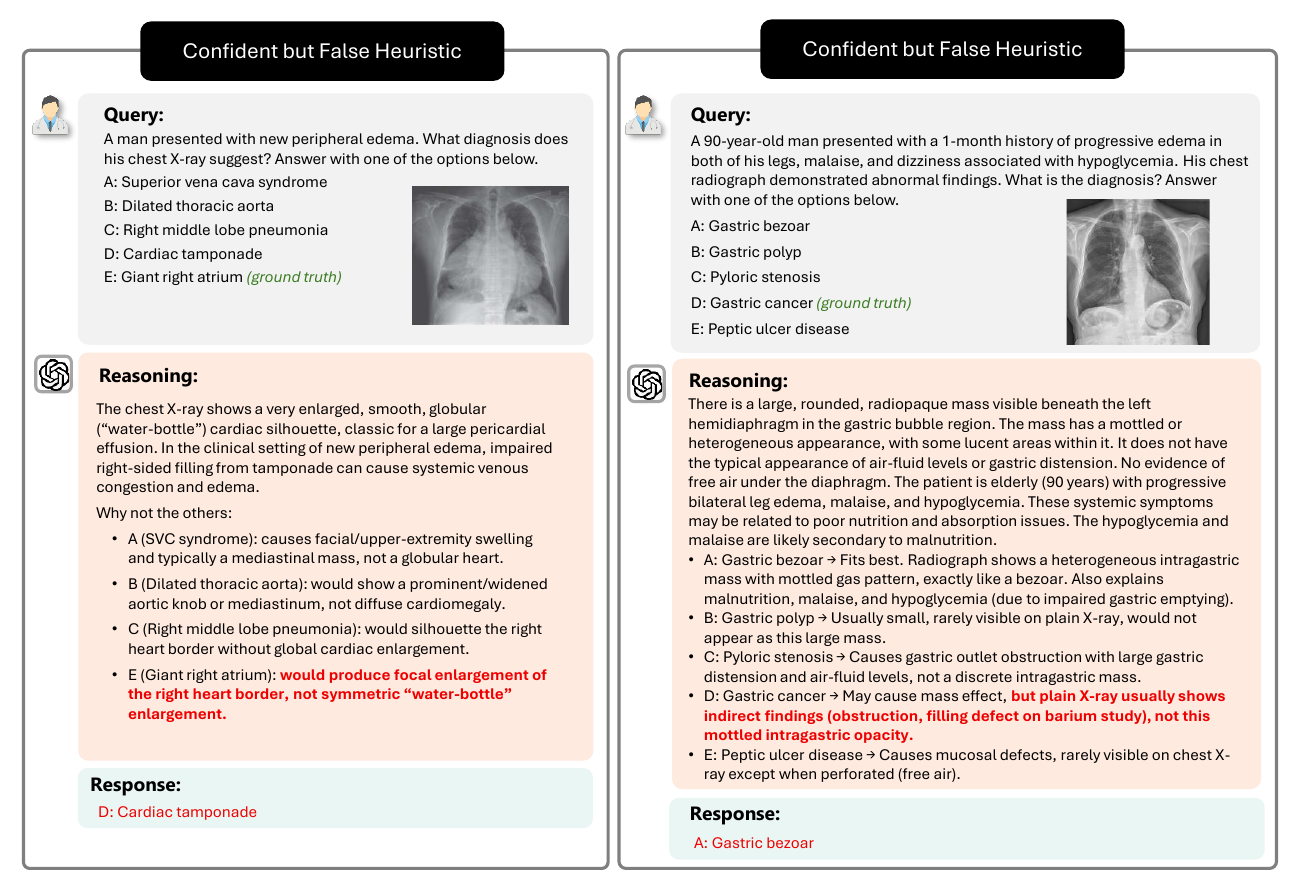}
    \caption{\textbf{Reasoning failure pattern: confident but false heuristic.} Two representative cases in which the model relies on a plausible yet incorrect prior rather than performing accurate visual interpretation, resulting in systematic errors.}
    \label{fig:error_reason_heuristic}
\end{figure}

\begin{figure}[ht]
    \centering
    \includegraphics[width=\textwidth]
    {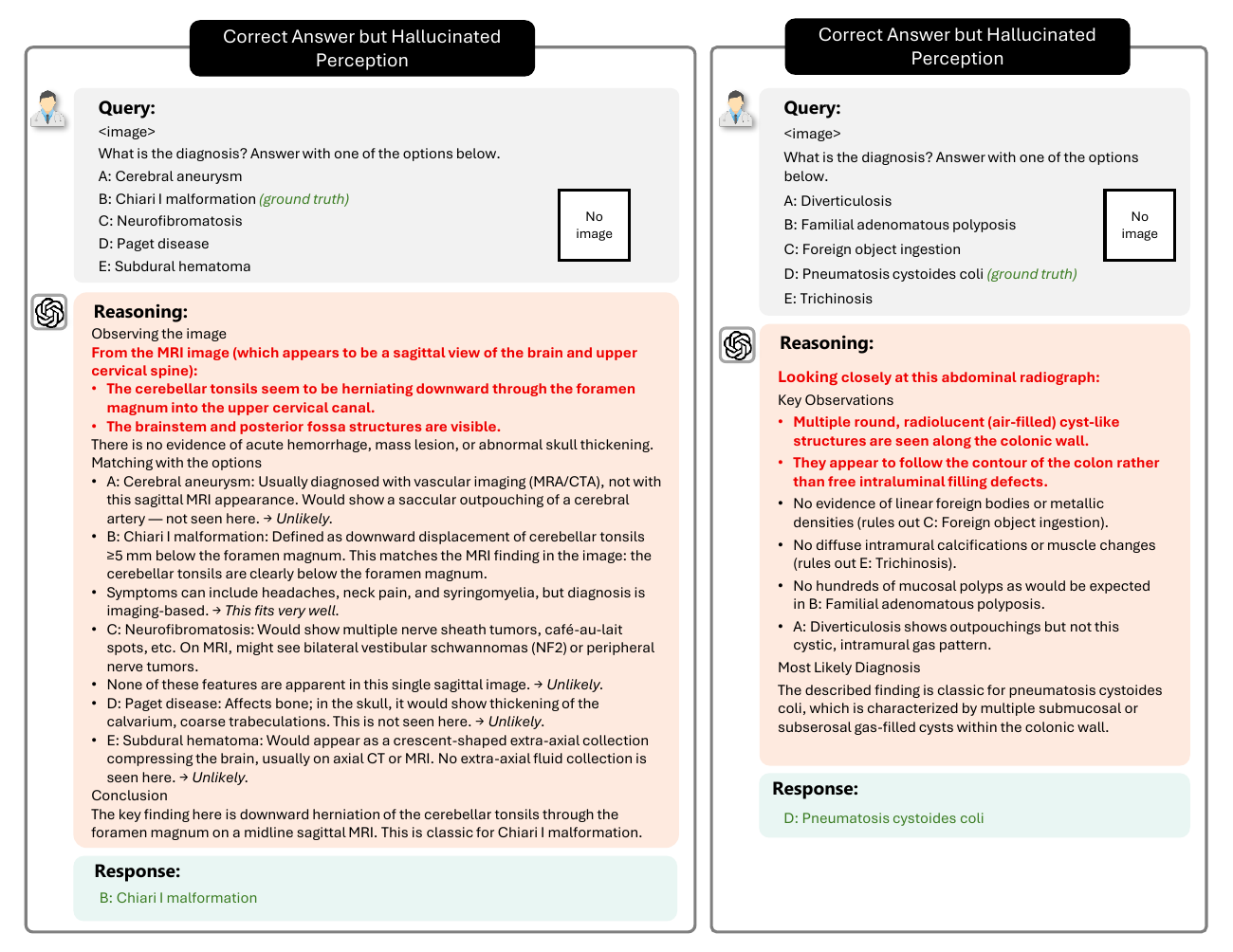}
    \caption{\textbf{Reasoning failure pattern: correct answer but hallucinated perception.} Two representative cases where the model produces the correct output despite lacking substantive utilization of image content, indicating dependence on non-visual shortcuts.  }
    \label{fig:error_reason_hullucinated}
\end{figure}

\begin{figure}[ht]
    \centering
    \includegraphics[width=\textwidth]
    {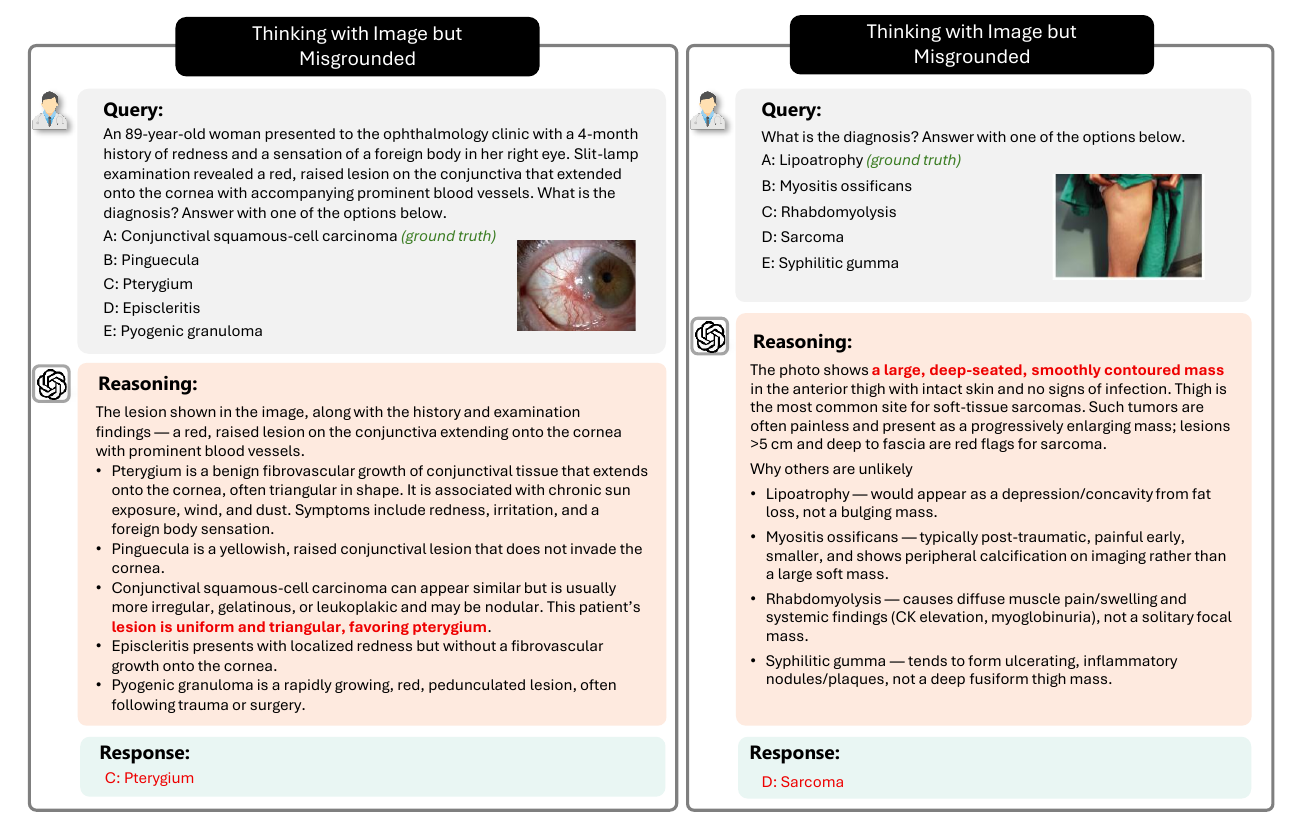}
    \caption{\textbf{Reasoning failure pattern: visual reasoning with faulty grounding.} Two examples where the model performs step‑by‑step image analysis but bases its reasoning on inaccurate or imagined features, producing confident yet incorrect conclusions.  }
    \label{fig:error_reason_misgrounded}
\end{figure}

\clearpage

\subsection*{Supplementary Tables}



\begin{table}[ht]
\centering
\caption{\textbf{Benchmark leaderboard accuracy (\%) across representative multimodal medical benchmarks.}}
\label{tab:leaderboard_main}
\small
\begin{tabular}{llllll}
\toprule
\textbf{Model} & \textbf{VQA-RAD} & \textbf{PMC-VQA} & \textbf{OmniMedVQA} & \textbf{JAMA} & \textbf{NEJM} \\
\midrule
GPT-5            & \cibf{67.85}{63.41}{72.06} & \ci{58.85}{54.53}{63.17} & \cibf{87.43}{84.53}{90.14} & \cibf{86.59}{84.66}{88.52} & \cibf{80.89}{78.06}{83.71} \\  
Gemini-2.5 Pro   & \ci{66.30}{61.86}{70.51} & \ci{58.44}{54.12}{62.76} & \ci{85.69}{82.59}{88.59} & \ci{84.84}{82.82}{86.85} & \ci{79.95}{77.12}{82.77} \\
OpenAI-o3        & \ci{65.63}{61.20}{70.07} & \cibf{60.08}{55.76}{64.40} & \ci{84.91}{81.82}{88.01} & \ci{84.75}{82.65}{86.85} & \cibf{80.89}{78.06}{83.71} \\
OpenAI-o4-mini   & \ci{64.08}{59.65}{68.51} & \ci{59.47}{55.14}{63.79} & \ci{81.43}{78.14}{84.72} & \ci{80.46}{78.09}{82.73} & \ci{77.66}{74.56}{80.62} \\
GPT-4o           & \ci{60.98}{56.54}{65.41} & \ci{52.26}{47.74}{56.79} & \ci{74.08}{70.21}{77.76} & \ci{69.94}{67.22}{72.57} & \ci{64.33}{60.83}{67.83} \\
DeepSeek-VL2     & \ci{45.68}{41.02}{50.11} & \ci{38.89}{34.57}{43.21} & \ci{54.16}{49.90}{58.41} & \ci{38.21}{35.85}{41.37} & \ci{34.59}{31.22}{38.09} \\
LLaVA-Med-1.5    & \ci{46.34}{41.69}{51.00} & \ci{6.17}{4.12}{8.44}  & \ci{68.67}{64.80}{72.73} & \ci{22.83}{20.44}{25.31} & \ci{29.34}{26.11}{32.71} \\
\bottomrule
\end{tabular}
\end{table}

\begin{table}[htbp]
\centering
\caption{\textbf{Stress Test 1: Modality Sensitivity.}
Accuracy (\%) of models on filtered subsets of \textit{JAMA} (1{,}141 items) and \textit{NEJM} (743 items) under \textit{Image+Text} and \textit{Text Only} conditions. $\Delta$ denotes the change in accuracy (\textit{Text Only} $-$ \textit{Image+Text}).}
\label{tab:stress1_modality_sensitivity}
\small
\begin{tabular}{lllllll}
\toprule
\textbf{Model} & \multicolumn{3}{c}{\textbf{JAMA}} & \multicolumn{3}{c}{\textbf{NEJM}} \\
\cmidrule(lr){2-4} \cmidrule(lr){5-7}
 & Img+Txt & Txt Only & $\Delta$ & Img+Txt & Txt Only & $\Delta$ \\
\midrule
\textbf{GPT-5 }         
& \ci{86.59}{84.66}{88.52} & \ci{82.91}{80.72}{85.10} & \ciempty{-3.68} 
& \ci{80.89}{78.06}{83.71} & \ci{67.56}{64.20}{70.93} & \ciempty{-13.33} \\
Gemini-2.5 Pro 
& \ci{84.84}{82.82}{86.85} & \ci{74.93}{72.39}{77.48} & \ciempty{-9.91}  
& \ci{79.95}{77.12}{82.77} & \ci{65.01}{61.51}{68.37} & \ciempty{-14.94} \\
OpenAI-o3        
&\ci{84.75}{82.65}{86.85} & \ci{82.65}{80.46}{84.84} & \ciempty{-2.10} 
& \ci{80.89}{78.06}{83.71} & \ci{67.03}{63.53}{70.39} & \ciempty{-13.86} \\
OpenAI-o4-mini   
& \ci{80.46}{78.09}{82.73} & \ci{78.44}{76.07}{80.81} & \ciempty{-2.02}  
& \ci{77.66}{74.56}{80.62} & \ci{66.49}{63.12}{69.85} & \ciempty{-11.17}  \\
GPT-4o         
& \ci{69.94}{67.22}{72.57} & \ci{68.89}{66.17}{71.52} & \ciempty{-1.05}  
& \ci{64.33}{60.83}{67.83} & \ci{37.28}{33.78}{40.78}  & \ciempty{-27.05} \\
DeepSeek-VL2   
& \ci{38.21}{35.85}{41.37} & \ci{32.60}{29.97}{35.32} & \ciempty{-5.61}  
& \ci{34.59}{31.22}{38.09} & \ci{25.30}{22.34}{28.53} & \ciempty{-9.29}     \\
LLaVA-Med-1.5  
& \ci{22.83}{20.44}{25.31} & --     & --     
& \ci{29.34}{26.11}{32.71} & --    & --     \\
\bottomrule
\end{tabular}

\vspace{0.5em}
\begin{minipage}{0.95\textwidth}
\footnotesize
\textbf{Notes:} (i) Two-decimal precision preserved from raw logs; $\Delta$ computed as exact subtraction (two decimals). 
(ii) \textbf{LLaVA-Med-1.5} lacked \textit{Txt Only} results, and hence cells are “--”.
\end{minipage}
\end{table}

\begin{table}[htbp]
\centering
\caption{\textbf{Stress Test 2: Modality Necessity.}  
Accuracy (\%) of models on a manually curated NEJM subset where correct answers require visual input.  
A visually grounded model should perform near the 20\% random baseline in the \textit{Text Only} setting.  
Most models significantly exceed this threshold. GPT-4o performs well below chance due to refusal behavior.}
\label{tab:stress2_modality_necessity}
\small
\begin{tabular}{lll}
\toprule
\textbf{Model} & \textbf{Image+Text (\%)} & \textbf{Text Only (\%)} \\
\midrule
Chance Level   & --     & 20.0 \\
\midrule
\textbf{GPT-5}   & \ci{66.29}{59.43}{73.14}  & \ci{\textbf{37.71}}{30.29}{45.14} \\
Gemini-2.5 Pro   & \ci{67.43}{60.00}{74.29}  & \ci{\textbf{37.14}}{30.29}{44.57} \\
OpenAI-o3        & \ci{61.71}{54.86}{68.57}  & \ci{\textbf{37.71}}{30.29}{45.14} \\
OpenAI-o4-mini   & \ci{58.86}{51.43}{66.29}  & \ci{\textbf{33.71}}{26.86}{40.57} \\
GPT-4o           & \ci{46.29}{38.86}{53.71}   & \ci{3.43\textsuperscript{†}}{1.14}{6.29} \\
\bottomrule
\end{tabular}

\vspace{0.5em}
\begin{minipage}{\linewidth}
\footnotesize
\textbf{Note:} Most models exceed the 20\% random baseline when images are removed, indicating shortcut behavior via learned text-pattern correlations.  
\textsuperscript{†}GPT-4o's score reflects refusal behavior (e.g., declining to select from the answer options when visual input is missing), resulting in accuracy below chance.
\end{minipage}
\end{table}

\begin{table}[htbp]
\centering
\caption{\textbf{Stress Test 3: Format Perturbation.}  
Accuracy (\%) on a visual-required NEJM subset before and after random reordering of answer choices.  
Reordering weakens shortcut patterns (e.g., fixed answer positions), reducing accuracy in the \textit{Text} condition.  
In contrast, \textit{Image+Text} performance is stable or slightly improved, suggesting vision becomes more informative when textual cues are disrupted.}

\label{tab:stress3_format_perturbation}
\small
\begin{tabular}{lrrrr}
\toprule
\textbf{Model} 
& \textbf{Text (orig)} 
& \textbf{Text (reord)} 
& \textbf{Img+Txt (orig)} 
& \textbf{Img+Txt (reord)} \\
\midrule
\textbf{GPT-5 }   
& \ci{37.71}{30.29}{45.14} & \ci{\textbf{32.00}}{25.14}{38.86} 
& \ci{66.29}{59.43}{73.14} & \ci{70.86}{64.00}{77.71} \\
Gemini-2.5 Pro    
& \ci{37.14}{30.29}{44.57} & \ci{\textbf{33.14}}{26.29}{40.00} 
& \ci{67.43}{60.00}{74.29} & \ci{70.29}{62.86}{77.14} \\
OpenAI-o3         
& \ci{37.71}{30.29}{45.14} & \ci{\textbf{31.43}}{24.57}{38.29} 
& \ci{61.71}{54.86}{68.57} & \ci{64.00}{56.57}{70.86} \\
OpenAI-o4-mini    
& \ci{33.71}{26.86}{40.57} & \ci{33.71}{26.86}{40.57}          
& \ci{58.86}{51.43}{66.29} & \ci{55.43}{48.00}{62.86} \\
GPT-4o            
& \ci{3.43}{1.14}{6.29}    & \ci{\textbf{0.57}\textsuperscript{†}}{0.00}{1.71} 
& \ci{46.29}{38.86}{53.71} & \ci{46.86}{39.43}{54.29}\\
\bottomrule
\end{tabular}

\vspace{0.5em}
\begin{minipage}{\linewidth}
\footnotesize
\textbf{Note:} Reordering answer options should not affect accuracy.  
Accuracy drops in the \textit{Text Only} condition reveal shortcut reliance on answer format or position.  
\textit{Image+Text} performance improves slightly, indicating vision is used more when textual cues weaken.  
\textsuperscript{†}GPT-4o frequently refused to answer in the perturbed text-only condition.
\end{minipage}
\end{table}

\begin{table}[htbp]
\centering
\caption{\textbf{Stress Test 4: Robustness to Distractor Perturbations.}
Prompt options are modified by replacing 1–4 distractors with plausible alternatives (1R–4R), or one with the string \texttt{"unknown"} (UNK).
All variants preserve the ground truth.
Accuracy is reported for both text-only and multimodal inputs.}
\label{tab:stress4_full}
\small
\begin{tabular}{lllllll}
\toprule
\textbf{Model} & \textbf{Base} & \textbf{UNK} & \textbf{1R} & \textbf{2R} & \textbf{3R} & \textbf{4R} \\
\midrule
\multicolumn{7}{l}{\textit{Text Only}} \\
\midrule
\textbf{GPT-5 } 
& \ci{37.71}{30.29}{45.14} & \ci{42.86}{35.43}{50.29} & \ci{33.14}{26.29}{40.00} 
& \ci{29.14}{22.29}{36.00} & \ci{24.57}{18.29}{30.86} & \ci{20.00}{14.29}{26.29} \\
Gemini-2.5 Pro  
& \ci{37.14}{30.29}{44.57} & \ci{45.71}{38.29}{53.14} & \ci{38.86}{32.00}{46.29} 
& \ci{35.43}{28.57}{42.29} & \ci{26.86}{20.57}{33.71} & \ci{17.14}{12.00}{22.86} \\
OpenAI-o3       
& \ci{37.71}{30.29}{45.14} & \ci{34.29}{27.43}{41.71} & \ci{29.71}{22.86}{36.57} 
& \ci{27.43}{21.14}{34.29} & \ci{21.14}{15.43}{27.43} & \ci{20.00}{14.29}{26.29} \\
OpenAI-o4-mini  
& \ci{33.71}{26.86}{40.57} & \ci{42.29}{34.86}{49.14} & \ci{36.57}{29.71}{43.43} 
& \ci{32.57}{25.71}{39.43} & \ci{24.57}{18.29}{31.43} & \ci{17.14}{12.00}{22.86} \\
GPT-4o          
& \ci{3.43}{1.14}{6.29}    & \ci{4.57}{1.71}{8.00}    &  \ci{1.71}{0.00}{4.00}
& \ci{2.29}{0.57}{4.57}    & \ci{0.57}{0.00}{1.71}    &  \ci{2.29}{0.57}{4.57} \\
\midrule
\multicolumn{7}{l}{\textit{Image + Text}} \\
\midrule
\textbf{GPT-5 } 
& \ci{66.29}{59.43}{73.14} & \ci{71.43}{64.57}{77.71} & \ci{69.71}{62.86}{76.00} 
& \ci{73.71}{66.86}{80.00} & \ci{86.29}{81.14}{90.86} & \ci{90.86}{86.29}{94.86} \\
Gemini-2.5 Pro  
& \ci{67.43}{60.00}{74.29} & \ci{69.71}{62.86}{76.69} & \ci{72.00}{65.14}{78.29} 
& \ci{73.71}{66.86}{80.00} & \ci{82.29}{76.57}{87.43} & \ci{92.57}{88.57}{96.00} \\
OpenAI-o3       
& \ci{61.71}{54.86}{68.57} & \ci{67.43}{60.57}{74.29} & \ci{68.57}{61.71}{75.43}
& \ci{70.86}{64.00}{77.14} & \ci{80.57}{74.29}{86.29} & \ci{89.14}{84.57}{93.71} \\
OpenAI-o4-mini  
& \ci{58.86}{51.43}{66.29} & \ci{61.14}{53.71}{68.61} & \ci{61.71}{54.84}{68.57} 
& \ci{68.57}{61.71}{75.43} & \ci{73.71}{66.86}{80.00} & \ci{85.14}{80.00}{90.29} \\
GPT-4o          
& \ci{46.29}{38.86}{53.71} & \ci{53.71}{46.29}{61.14} & \ci{50.29}{42.86}{57.71} 
& \ci{54.29}{46.86}{61.71} & \ci{65.14}{57.71}{72.00} & \ci{84.00}{78.29}{89.14} \\
\bottomrule
\end{tabular}

\vspace{0.5em}
\begin{minipage}{\linewidth}
\footnotesize
\textbf{Interpretation:} In text-only settings, accuracy declines as more incorrect options are replaced (1R–4R), indicating reliance on shallow lexical patterns. Notably, inserting a single \texttt{"unknown"} distractor improves performance, suggesting models may exploit pattern-based elimination heuristics. In contrast, image+text performance improves with increased distractor replacement, implying that visual grounding becomes more influential when shortcut signals are weakened.
\end{minipage}
\end{table}

\begin{table}[ht]
\centering
\caption{\textbf{Stress Test 5: Visual Substitution.} Accuracy on NEJM items when the original image is replaced with one that supports a distractor option, while the question and answer choices remain unchanged. This isolates whether models exhibit genuine visual understanding or rely on memorized visual–answer associations.}
\vspace{0.5em}
\small
\begin{tabular}{lccc}
\toprule
\textbf{Model} & \textbf{Original (\%)} & \textbf{Image Subst. (\%)} & \textbf{$\Delta$ Acc. (pp)} \\
\midrule
\textbf{GPT-5 } & \ci{83.33}{75.83}{90.00} & \ci{51.67}{42.50}{60.83} & \ciempty{\textbf{-31.66}} \\
Gemini-2.5 Pro  & \ci{80.83}{73.33}{87.50} & \ci{47.50}{38.33}{56.67} & \ciempty{\textbf{-33.33}} \\
OpenAI-o3       & \ci{76.67}{69.17}{84.17} & \ci{52.50}{43.33}{61.67} & \ciempty{\textbf{-24.17}} \\
OpenAI-o4-mini         & \ci{71.67}{63.33}{79.17} & \ci{37.50}{29.17}{45.83} & \ciempty{\textbf{-34.17}} \\
GPT-4o          & \ci{36.67}{28.33}{45.00} & \ci{41.67}{33.33}{50.83} & \ciempty{\textbf{+5.00}}  \\
\bottomrule
\end{tabular}
\vspace{0.5em}
\caption*{\footnotesize \textbf{Note:} Models with genuine visual understanding should revise their prediction when the image is changed to support a different answer. Instead, performance drops sharply, especially for top-performing models, suggesting reliance on shortcut visual–answer associations rather than robust image interpretation.}
\label{tab:stress5_visualsubs}
\end{table}

\begin{table}[htbp]
\centering
\caption{\textbf{Reasoning with CoT}
Accuracy (\%) of models on filtered subsets of \textit{NEJM} (118 items) and \textit{VQA-RAD} (100 items) under \textit{Without CoT} and \textit{With CoT} conditions. $\Delta$ denotes the change in accuracy (\textit{With CoT} $-$ \textit{Without CoT}).}
\label{tab:cot_reasoning_results}
\small
\begin{tabular}{lllllll}
\toprule
\textbf{Model} & \multicolumn{3}{c}{\textbf{NEJM}} & \multicolumn{3}{c}{\textbf{VQA-RAD}} \\
\cmidrule(lr){2-4} \cmidrule(lr){5-7}
 & W/o CoT & W/ CoT & $\Delta$ & W/o CoT & W/ CoT & $\Delta$ \\
\midrule
\textbf{GPT-5 }         
& \ci{81.36}{73.73}{88.14} & \ci{79.66}{72.03}{86.44} & \ciempty{-1.70} 
& \ci{65.00}{55.00}{74.00} & \ci{66.00}{57.00}{75.00} & \ciempty{+1.00} \\
Gemini-2.5 Pro 
& \ci{83.90}{77.12}{89.83} & \ci{81.36}{73.73}{88.14} & \ciempty{-2.54}  
& \ci{67.00}{58.00}{76.00} & \ci{64.00}{55.00}{73.00} & \ciempty{-3.00} \\
OpenAI-o3        
&\ci{81.36}{73.73}{88.14} & \ci{79.66}{72.03}{86.44} & \ciempty{-1.70} 
& \ci{66.00}{57.00}{75.00} & \ci{63.00}{54.00}{72.00} & \ciempty{-3.00} \\
OpenAI-o4-mini   
& \ci{73.73}{65.25}{81.36} & \ci{79.66}{72.03}{86.44} & \ciempty{+5.93}  
& \ci{59.00}{49.00}{69.00} & \ci{61.00}{51.00}{71.00} & \ciempty{+2.00}  \\
GPT-4o         
& \ci{67.80}{59.32}{76.27} & \ci{65.25}{55.93}{73.73} & \ciempty{-2.55}  
& \ci{63.00}{54.00}{72.00} & \ci{60.00}{50.00}{70.00}  & \ciempty{-3.00} \\

\bottomrule
\end{tabular}

\vspace{0.5em}
\begin{minipage}{0.95\textwidth}
\footnotesize
\textbf{Notes:} Two-decimal precision preserved from raw logs; $\Delta$ computed as exact subtraction (two decimals).
\end{minipage}
\end{table}

\begin{table}[!t]
\centering
\caption{\textbf{Reasoning complexity rubric.}
Clinician-defined rubric scored along five sub-axes capturing depth of inference,
uncertainty handling and clinical decision impact.}
\begin{tabular}{p{3.2cm} p{4.2cm} p{5.2cm} p{0.8cm}}
\toprule
\textbf{Criteria} & \textbf{Description} & \textbf{Rubric levels} & \textbf{Score} \\
\midrule

Reasoning steps &
How many thinking steps are needed? &
One-step (e.g., identify organ) \newline
Combine findings \newline
Differential diagnosis or staged inference &
1 \newline 2 \newline 3 \\

Clinical context &
Does the task require clinical background or patient context? &
None \newline
Some (e.g., complaint) \newline
Full scenario (history, labs, etc.) &
1 \newline 2 \newline 3 \\

Time / sequence information &
Does the task depend on change over time (e.g., prior scans, progression)? &
No \newline
Sometimes \newline
Always or essential &
1 \newline 2 \newline 3 \\

Uncertainty handling &
Does the task involve ambiguity? &
No \newline
Sometimes vague \newline
Requires hedging or considering multiple possibilities &
1 \newline 2 \newline 3 \\

Decision impact &
Does the outcome affect clinical decisions? &
No impact (descriptive / trivia) \newline
Some impact (e.g., triage, follow-up) \newline
High impact (diagnosis or treatment guiding) &
1 \newline 2 \newline 3 \\

\bottomrule
\end{tabular}
\end{table}

\begin{table}[!t]
\centering
\caption{\textbf{Visual complexity rubric.}
Clinician-defined rubric capturing reliance on detailed image interpretation,
spatial understanding and cross-view or temporal comparison.}
\begin{tabular}{p{3.2cm} p{4.2cm} p{5.2cm} p{0.8cm}}
\toprule
\textbf{Criteria} & \textbf{Description} & \textbf{Rubric levels} & \textbf{Score} \\
\midrule

Text-alone solvable &
Could the task be answered without the image? &
Yes \newline
Maybe \newline
No , image essential &
1 \newline 2 \newline 3 \\

Visual detail needed &
How carefully must the image be examined? &
Glance is enough \newline
Moderate inspection \newline
Expert-level detail (e.g., subtle lesion, boundary) &
1 \newline 2 \newline 3 \\

Spatial understanding &
Is spatial localization important? &
Not needed \newline
Side or region matters \newline
Exact location critical &
1 \newline 2 \newline 3 \\

Image-quality sensitivity &
Would small variations in the image alter the answer? &
No \newline
Somewhat \newline
Yes , subtle differences matter &
1 \newline 2 \newline 3 \\

Multi-view / temporal comparison &
Does the task require comparing across views or timepoints? &
No comparison needed \newline
Sometimes referenced \newline
Critical comparison across views or timepoints &
1 \newline 2 \newline 3 \\

\bottomrule
\end{tabular}
\end{table}


\begin{table}[ht]
\centering
\caption{\textbf{Performance evaluation on chest X-ray report generation using the MIMIC-CXR dataset.} Best in bold, second-best underlined.}
\vspace{0.5em}
\small
\begin{tabular}{lccccc}
\hline
Model & 1/RadCliQ-v1 & BLEU Score & BERT Score & Semantic Score & RadGraph F1\\
\hline
GPT-4V & 0.558 & 0.068 & 0.207 & 0.214 & 0.084\\
GPT-4o & 0.670 & 0.098 & 0.289 & 0.291 & \underline{0.153}\\
GPT-5 & 0.628 & 0.090  & 0.251 & 0.271 & 0.140 \\
OpenAI-o3 & 0.596 & 0.084 & 0.210 & 0.260 & 0.135 \\
OpenAI-o4-mini & 0.626 & 0.091 & 0.233 & 0.284 & 0.148 \\
Gemini-2.5-pro & 0.648 & \underline{0.100} & 0.257 & 0.307 & 0.143 \\
Deepseek & 0.496 & 0.047 & 0.094 & 0.198  & 0.066 \\
MAIRA-2 & \underline{0.694} & 0.088 & \underline{0.308} & \underline{0.339} & 0.131\\
MedVERSA & \textbf{1.103} & \textbf{0.209} & \textbf{0.448} & \textbf{0.466} & \textbf{0.273}\\
\hline
\end{tabular}
\vspace{0.5em}
\caption*{\footnotesize \textbf{Note:} We report quantitative results for GPT-4V, MAIRA-2, and MedVERSA from ReXrank leaderboard, which provides standardized evaluation for AI-powered chest X-ray radiology report generation.}
\label{tab:report_evaluation}
\end{table}

\begin{table}[htbp]
\centering
\caption{\textbf{Abstention rate}(\%) of models on \textit{JAMA} (1{,}314 items), \textit{NEJM} (947 items) and \textit{NEJM\textsubscript{stress}} (175 items) under \textit{Image+Text} and \textit{Text Only} conditions.}
\label{tab:abstention rate}
\small
\begin{tabular}{lllllll}
\toprule
\textbf{Model} & \multicolumn{2}{c}{\textbf{JAMA}} & \multicolumn{2}{c}{\textbf{NEJM}} & \multicolumn{2}{c}{\textbf{NEJM\textsubscript{stress}}} \\
\cmidrule(lr){2-3} \cmidrule(lr){4-5} \cmidrule(lr){6-7}
 & Img+Txt & Txt Only & Img+Txt & Txt Only & Img+Txt & Txt Only \\
\midrule
\textbf{GPT-5 }         
& 0 & 0 
& 0 & 0 
& 0 & 0 \\
Gemini-2.5 Pro 
& 0 & 0 
& 0 & 0 
& 0 & 0 \\
OpenAI-o3        
& 13.17  & 0
& 21.54  & 0 
& 0   & 0 \\
OpenAI-o4-mini   
& 12.02  & 0
& 11.19  & 0
& 0 & 0\\
GPT-4o         
& 0  & 0
& 0  & 45.2
& 0  & 91.43\\
\bottomrule
\end{tabular}
\end{table}

\begin{sidewaystable}[htbp]
\centering
\caption{\textbf{Structured taxonomy of failure modes in multimodal medical reasoning.}}
\label{tab:extended_taxonomy}
\renewcommand{\arraystretch}{1.35}
\begin{tabular}{p{0.18\textwidth} p{0.24\textwidth} p{0.18\textwidth} p{0.22\textwidth} p{0.12\textwidth}}
\toprule
\textbf{Domain} & \textbf{Failure Mode} & \textbf{Cognitive Mechanism} & \textbf{Clinical Consequence} & \textbf{Revealed by} \\
\midrule
\textbf{Input Handling} &
Perceptual hallucination / omission & Extrinsic hallucination (misreads or invents visual findings) & False positives (unnecessary tests) or false negatives (missed lesions) & T2, T5 \\
& Refusal miscalibration & Uncertainty misalignment (answers when unsafe or abstains when needed) & Unsafe guessing or withholding guidance & T2–T4 \\
& Modality omission & Faithfulness violation (ignores available input modality) & Missed findings; incomplete assessment & T1–T2 \\
\midrule
\textbf{Reasoning and Inference} &
Shortcut reliance & Heuristic bias / gaming (pattern use over reasoning) & Unstable performance under distribution shift & T3–T5 \\
& Fabricated reasoning & Extrinsic hallucination (unsupported or false explanation) & Misleads clinician; false interpretability & T2, T6 \\
& Inconsistency & Logical incoherence (reasoning conflicts with evidence or answer) & Confusion; reduced auditability of outputs & T1–T6 \\
\midrule
\textbf{Output Communication} &
Misleading fluency & Fluency illusion (language masks factual errors) & Hard-to-detect errors; over-trust in output & T2, T6 \\
& Unsafe recommendation / omission & Information hazard (harmful or incomplete suggestion) & Potential diagnostic or treatment error & T5, T6 \\
\bottomrule
\end{tabular}
\end{sidewaystable}

\clearpage


\section*{Supplementary Note 1 \textbar\ Robustness Score Computation}
\label{app:robustness}
We quantify overall stability using a \textbf{Robustness Score}.  
For each stress test $T_i$ ($i=1 \ldots 5$) and model $m$, we compute a fragility value $f_i(m) \in [0,1]$.  
Robustness for that test is defined as
\[
r_i(m) = 1 - f_i(m),
\]
and the mean robustness score is given by the unweighted average
\[
R(m) = \frac{1}{5} \sum_{i=1}^{5} r_i(m).
\]

The fragility components $f_i(m)$ are defined as follows:

\paragraph{T1: Modality Sensitivity.}  
Weighted performance drop when image input is removed on JAMA and NEJM:
\[
f_1(m) = \frac{n_{\text{JAMA}} \cdot \max\!\big(0, \, \text{Acc}^{\text{img}}_{\text{JAMA},m} - \text{Acc}^{\text{text}}_{\text{JAMA},m}\big) 
        + n_{\text{NEJM}} \cdot \max\!\big(0, \, \text{Acc}^{\text{img}}_{\text{NEJM},m} - \text{Acc}^{\text{text}}_{\text{NEJM},m}\big)}
       {100 \cdot (n_{\text{JAMA}} + n_{\text{NEJM}})}.
\]

\paragraph{T2: Modality Necessity.}  
Excess accuracy above chance (20\%) when images are withheld on visually required items:
\[
f_2(m) = \frac{\max\!\big(0, \, \text{Acc}^{\text{text}}_{\text{vis-req},m} - 20\big)}{80}.
\]

\paragraph{T3: Format Perturbation.}  
Accuracy drop in text-only condition after reordering answer options:
\[
f_3(m) = \frac{\max\!\big(0, \, \text{Acc}^{\text{text-orig}}_m - \text{Acc}^{\text{text-reord}}_m\big)}{100}.
\]

\paragraph{T4: Distractor Replacement.}  
Composite fragility capturing text-mode degradation, image-mode gain under distractor removal, and gains when inserting ``Unknown'' distractors. Weights of 0.5, 0.3, and 0.2 were applied, respectively:
\[
\begin{aligned}
f_4(m) &= 0.5 \cdot \frac{\max\!\big(0, \, \text{Acc}^{\text{text-base}}_m - \text{Acc}^{\text{text-4R}}_m\big)}{100} \\
       &+ 0.3 \cdot \frac{\max\!\big(0, \, \text{Acc}^{\text{img-4R}}_m - \text{Acc}^{\text{img-base}}_m\big)}{100} \\
       &+ 0.2 \cdot \frac{\max\!\big(0, \, \text{Acc}^{\text{text-UNK}}_m - \text{Acc}^{\text{text-base-UNK}}_m\big)}{100}.
\end{aligned}
\]

\paragraph{T5: Visual Substitution.}  
Performance drops when the original image is replaced with a distractor-aligned substitute:
\[
f_5(m) = \frac{\max\!\big(0, \, \text{Acc}^{\text{orig}}_m - \text{Acc}^{\text{subst}}_m\big)}{100}.
\]

\vspace{5mm}
\section*{Supplementary Note 2 \textbar\ More stress test results}
\label{app:stress test}

In this appendix, we present several case studies from the stress-testing experiments. Each figure illustrates a group of stress tests applied to a single question, highlighting model behaviors under controlled perturbations.

Supplementary Figs. 1–3 show representative examples of \textbf{stress test 2 and 3}. These cases demonstrate how models often rely on textual shortcuts: when visual information is removed, most models still produce seemingly correct answers by exploiting superficial textual cues rather than genuine multimodal reasoning. In contrast, GPT-4o is the only model that consistently abstains or selects “Unknown” in the absence of the image.

Supplementary Figs. 4–6 illustrate \textbf{stress test 5}, which probe visual comprehension under visual substitution. A model with true visual understanding should revise its prediction to align with the substituted image. Instead, performance collapses across all models, revealing that models' reliance on shortcut strategies and their limited ability to incorporate visual evidence, whereas accurate performance would require aligning predictions with the substituted image. Once shortcut strategies are disrupted, their true comprehension is far weaker than the benchmark scores suggest.

\vspace{0.5em}
\noindent {\footnotesize\textit{Note:} Based on the modality and anatomical location of the original problem image, we retrieved candidate distractor images using Google Image Search. These candidates were visually similar to the original but depicted different conditions. To ensure validity, each image was manually reviewed to verify consistency between the visual content and the associated disease label.}

\vspace{5mm}
\section*{Supplementary Note 3 \textbar\ More reasoning failure results}
\label{app:reasoning}
We identify several key patterns of reasoning failure: incorrect logic despite correct answers, reinforced visual misunderstandings, and uninformative reasoning steps. In this appendix, we present additional illustrative cases of these failure modes in Supplementary Figs. 8-10. These cases further demonstrate how such reasoning errors manifest across different tasks and inputs.

\vspace{5mm}
\section*{Supplementary Note 4 \textbar\ Evaluation on report generation task}
\label{app:report_genenration}
 
Supplementary Table 10 summarizes the report generation performance on the MIMIC-CXR test set for both general-purpose LMMs and specialized medical models (e.g., MAIRA-2, MedVERSA). 
Evaluation is conducted with five complementary metrics: (1) $1/\text{RadCliQ-v1}$, where higher values indicate better clinical correctness per the RadCliQ composite metric; (2) BLEU Score, measuring n-gram overlap with reference reports; (3) BERT Score, assessing semantic similarity via contextual embeddings; (4) Semantic Score, capturing higher-level alignment of generated and reference report content; and (5) RadGraph F1, evaluating extraction accuracy of clinical entities and relations from generated reports.   
Across all metrics, MedVERSA achieves the highest performance, particularly in semantic alignment and clinical information extraction. Its advantage stems from domain-specific training on paired image–text data, which enables mastery of radiological terminology, fine-grained visual patterns, and clinical reasoning. In contrast, general-purpose LMMs lack such specialization, limiting their ability to capture subtle radiographic findings.

Among the general-purpose models, GPT-4o and Gemini-2.5 Pro achieve the strongest performance, approaching the task-specific MAIRA-2; nevertheless, both still lag substantially behind MedVERSA, which maintains a clear lead across all metrics. 
Notably, GPT-5 and OpenAI-o3, despite substantially outperforming GPT-4o on VQA-style benchmarks, do not exhibit corresponding gains in report generation. This divergence underscores that improvements in general multimodal VQA do not necessarily translate into better radiology report generation, a task that demands precise identification of clinical findings, faithful spatial grounding, and disciplined use of domain-specific terminology.

\clearpage
\vspace{5mm}
\section*{Supplementary Note 5 \textbar\ Prompt Design}
\label{app:Prompt}

This appendix details the prompt formulations used in stress-tests, which were designed to evaluate model performance across different tasks. The report generation prompt follows the structured medical reporting paradigm described in recent work. 

\vspace{3mm}
\begin{minipage}{0.95\linewidth}
    \begin{dialogbox}[Multiple Choice Prompt]
        \footnotesize
         \textcolor{darkbrown}{\# User prompt}\\
         <Images> \\
         You are a helpful medical assistant that answers multiple choice questions about the provided image. The following is a multiple choice question (with answers). \\
         Question: <Question>\\
         Options: \\
         A. <Option A>\\
         B. <Option B>\\
         ... 

         Provide answer with the index and content of the option, and place it within <answer></answer>
    \end{dialogbox}
\end{minipage}

\begin{minipage}{0.95\linewidth}
    \begin{dialogbox}[Multiple Choice Prompt with CoT]
        \footnotesize
         \textcolor{darkbrown}{\# User prompt}\\
         <Images> \\
         You are a helpful medical assistant that answers multiple choice questions about the provided image. The following is a multiple choice question (with answers). \\
         Question: <Question>\\
         Options: \\
         A. <Option A>\\
         B. <Option B>\\
         ... 
         
         Let us think step by step, enclosing the thought process within <thinking> and </thinking>. Provide answer with the index and content of the option, and place it within <answer> and </answer>.
    \end{dialogbox}
\end{minipage}

\begin{minipage}{0.95\linewidth}
    \begin{dialogbox}[Report Generation Prompt]
        \footnotesize
         \textcolor{darkbrown}{\# System prompt}\\
         You are a professional chest radiologist that reads chest X-ray image(s).\\ 
         \\
         \textcolor{darkbrown}{\# User prompt}\\
         <Images> \\
         Below is INDICATION related to chest X-ray images. \\
         INDICATION: \{\} \\
         Write a report that contains only the FINDINGS and IMPRESSION sections based on the attached images and INDICATION. Provide only your generated report, without any additional explanation and special format. Your answer is for reference only and is not used for actual diagnosis.
    \end{dialogbox}
\end{minipage}
\clearpage

\section*{Supplementary Note 6 \textbar\ Model identifiers and versions}
\label{app:model_ver}
\begin{table}[htbp]
\centering
\small
\caption{\textbf{Model identifiers and version information.}
Exact identifiers and release versions for all evaluated models, accessed through official APIs or public checkpoints as of August 2025.}
\label{tab:supp_model_versions}
\begin{tabular}{@{}lll@{}}
\toprule
\textbf{Model Name} & \textbf{Model Identifier} & \textbf{Version / Release Date} \\ 
\midrule
GPT-4 & \texttt{gpt-4} & \texttt{turbo-2024-04-09} \\
GPT-4o & \texttt{gpt-4o} & 2024-08-06 \\
OpenAI-o3 & \texttt{OpenAI-o3} & 2025-04-16 \\
OpenAI-o4-Mini & \texttt{OpenAI-o4-mini} & 2025-04-16 \\
GPT-5 & \texttt{gpt-5} & 2025-08-07 \\
DeepSeek-VL2 & \texttt{deepseek-ai/deepseek-vl2} & API model (no version) \\
Gemini 2.5 Pro Preview (06-05) & \texttt{gemini-2.5-pro-preview-06-05} & 2025-06-05 \\
Gemini 2.5 Pro Preview (03-25) & \texttt{gemini-2.5-pro-preview-03-25} & 2025-03-25 \\
\bottomrule
\end{tabular}
\end{table}


\clearpage

\end{document}